\documentclass[12pt]{article}
\usepackage[utf8]{inputenc}
\usepackage{graphicx}
\usepackage{amsmath}
\usepackage{amsthm}
\usepackage{amssymb}
\usepackage{textcomp}
\usepackage{siunitx}
\usepackage{subfigure}
\usepackage{titling}
\usepackage{xcolor}
\usepackage{afterpage}
\usepackage{algorithm} 
\usepackage{algpseudocode} 
\usepackage{multirow}
\usepackage{hyperref}
\usepackage{algpseudocode}
\usepackage{mathtools}
\usepackage[normalem]{ulem}
\hypersetup{
    colorlinks=true,
    linkcolor=blue,
    citecolor = brown,
    filecolor=magenta,      
    urlcolor=cyan,
    pdftitle={Overleaf Example},
    pdfpagemode=FullScreen,
    }
\usepackage[font=small,labelfont=bf]{caption} % Required for specifying captions to tables and figures
\usepackage{graphicx} % Allows including images
\usepackage{booktabs}
\usepackage{geometry}
\newtheorem{definition}{Definition}
\newtheorem{proposition}{Proposition}[section]
\newtheorem{corollary}{Corollary}[section]

\newcommand{\R}{\mathbb{R}}

\newcommand{\N}{\mathbb{N}}

\newcommand{\E}{\mathbb{E}}

\newcommand{\argmax}{\text{argmax}}
\newcommand{\argmin}{\text{argmin}}
\newcommand{\cov}{\text{cov}}

\newcommand{\sign}{\textrm{sign}}

\providecommand{\keywords}[1]
{
  \small	
  \textbf{\textit{Keywords---}} #1
}

\title{Bayesian Autoregressive Online Change-Point Detection \\
with Time-Varying Parameters}
\author{Ioanna-Yvonni Tsaknaki\thanks{Universit\`a di Siena, Siena, Italy and Scuola Normale Superiore, Pisa, Italy. Email address: ioannayvonni.tsaknaki@sns.it} \and Fabrizio~Lillo\thanks{Dipartimento di Matematica, Universit\`a di Bologna and Scuola Normale Superiore, Pisa, Italy. Email address: fabrizio.lillo@unibo.it} \and Piero Mazzarisi\thanks{Universit\`a di Siena, Siena, Italy. Email address: piero.mazzarisi@unisi.it}}
\date{}

\begin{document}

\maketitle

\begin{abstract}
Change points in real-world systems mark significant regime shifts in system dynamics, possibly triggered by exogenous or endogenous factors. These points define regimes for the time evolution of the system and are crucial for understanding transitions in financial, economic, social, environmental, and technological contexts. Building upon the Bayesian approach introduced in \cite{c:07}, we devise a new method for online change point detection in the mean of a univariate time series, which is well suited for real-time applications and is able to handle the general temporal patterns displayed by data in many empirical contexts. We first describe time series as an autoregressive process of an arbitrary order. Second, the variance and correlation of the data are allowed to vary within each regime driven by a scoring rule that updates the value of the parameters for a better fit of the observations. Finally, a change point is detected in a probabilistic framework via the posterior distribution of the current regime length. By modeling temporal dependencies and time-varying parameters, the proposed approach enhances both the estimate accuracy and the forecasting power. Empirical validations using various datasets demonstrate the method's effectiveness in capturing memory and dynamic patterns, offering deeper insights into the non-stationary dynamics of real-world systems.
\end{abstract}


\section{Introduction}
Change Points (CPs) in real-world systems denote pivotal moments when the underlying characteristics and behaviors of these systems experience substantial and often abrupt alterations. These shifts can occur in various contexts, including financial, economic, social, environmental, and technological systems, to name but a few. CPs are typically triggered by events such as policy changes, natural disasters, technological breakthroughs, or social upheavals. For instance, in financial systems, a change point might be identified during a market crash or boom, reflecting sudden changes in investor behavior and economic conditions. Moreover, financial metrics such as price and volume result from the aggregation of individual trading behaviors at the micro level. Changes in these behaviors or the arrival of new investors can cause abrupt shifts in the overall market dynamics. In economic systems, change points mark transitions from periods of growth and stability to downturns characterized by declining GDP, rising unemployment, and reduced consumer spending. Such recessions can be triggered by various factors such as financial crises, abrupt changes in fiscal or monetary policies, geopolitical events, or significant disruptions in global trade. In environmental systems, change points might coincide with significant external events, such as the onset of prolonged droughts or sudden increases in global temperatures, leading to drastic shifts in ecosystem dynamics. In social systems, events such as political revolutions or major legislative changes can serve as change points that fundamentally alter societal structures and norms. The detection and analysis of these CPs are essential for researchers, policymakers, and stakeholders, as they provide critical insights into the timing and nature of these transitions.  Moreover, the ability to detect and understand these CPs enables a deeper understanding and aids in the development of strategies to manage change and address emerging challenges. Advanced analytical techniques in CP detection offer valuable tools for identifying these pivotal moments, thus enhancing the ability to navigate and adapt to the dynamic nature of real-world systems.

When a system can be described as a time series, Change-Point Detection (CPD) focuses on identifying points in time where the statistical properties of a time series change abruptly. These change points mark transitions between different states or regimes, indicating shifts in the underlying data-generating process. %The ability to accurately detect change points is critical for various applications. 
Traditional methods, such as likelihood ratio tests and cumulative sum procedures, provide robust frameworks for detecting single or multiple change points in relatively simple time series. However, as data have grown in size and complexity, more sophisticated approaches have been developed. Markov switching models, introduced in the milestone work of Hamilton \cite{Hamilton1988}, provide a framework for handling time-dependent data that exhibit change points in the parameters of an autoregressive model. These models operate by switching among a finite number of states. In this context, the transition times between regimes are treated as point estimates without any measure of uncertainty. Additionally, the assumption of a finite number of states imposes a significant restriction, as the parameters can only take on a limited set of values. Bayesian methods \cite{r:92,c:07}, on the contrary, offer a probabilistic framework that incorporates prior information and provides a comprehensive estimation of change points, parameters, and their uncertainties \cite{r:17}. Moreover, Bayesian inference allows for online updates of parameters whenever a new data point is observed, making it well-suited for real-time (or online) applications. However, the  feasibility of the Bayesian approach requires the specification of the data distribution (belonging to specific classes in order to simplify computations), typically assuming constant parameters within each regime. This is one of the reasons why nonparametric techniques \cite{brodsky}, which do not assume specific distributional forms, have also gained popularity for their flexibility in handling diverse types of data. Nevertheless, temporal dependencies are typically difficult to handle in nonparametric approaches.

In this paper, we build upon the seminal work of \cite{c:07} to propose a general Bayesian methodology for online change point detection in the mean of the distribution generating the time series. Our approach accounts for both temporal correlations in data and dynamic patterns in other moments of the distribution. Our method retains the benefits of uncertainty estimation within a probabilistic framework and allows for online parameter updates. To address non-Markovian dependencies, we describe the conditional probability of data with general memory order of past observations in closed form. A change point is inferred based on the run length, which represents the length of a regime. The probability distribution of this run length is estimated and updated with each new observation. Additionally, dynamic patterns in other moments of the data distribution, such as variance and (auto)correlation, are modeled using time-varying parameters driven by score updates as described in the general class of Score-Driven models of \cite{Score-Driven1}. These innovation terms adjust the parameters to better fit the observed data, similar to how GARCH models capture heteroskedastic effects in time series, with no prior assumption on the distribution of the time-varying parameters.

We corroborate the newly proposed methodology for CPD using three empirical datasets and show that memory and dynamic patterns in data matter. We further validate the results with a forecasting study. We start with the classical dancing bees dataset \cite{DancingBeesDataset}, which consists of the spatial trajectories of bees performing the waggle dance. Here we have a ground truth for the actual regime, so we can validate the performance of the proposed methodology using the actual CPs identified by specialists. These CPs include the movements of the waggle, turning left, and turning right, which are crucial for successful sharing of information about the direction and distance to the patches of flowers that produce nectar and pollen, water sources, or new nest site locations with other members of the colony. In all the empirical applications, we demonstrate that accounting for temporal dependencies memory and time-varying variance and correlation is crucial for accurately estimating change points and forecasting in the presence of regimes in time series. The second empirical application is to the problem of detecting growth and recessions periods in U.S. GNP data. Performing a comparative analysis with the pioneering study by \cite{Hamilton}, we emphasize that not all regimes associated with the same state of a Markov switching model are identical. Instead, growth periods and recessions exhibit different rates and intrinsic characteristics, leading to nonstationary patterns in the mean of the regimes, as captured by the proposed methodology. Finally, we complement the analysis of CPs for the dynamics of order flow in the financial market proposed in a companion paper \cite{QFpaper}. Since regimes in the time series of aggregated trading volumes likely indicate the execution of large trading programs in the market, we demonstrate how accounting for this information by using the proposed methodology aids in forecasting order flows. This capability is crucial for any trader aiming to minimize the transaction costs of their own orders.

The rest of the paper is organized as follows. Section \ref{sec:lit} reviews the scientific literature on CPD. Section \ref{sec:methods} reviews the baseline model of \cite{c:07} and describes the proposed change point detection methodology. Section \ref{sec:results} and Section \ref{sec:empirical} shows the Monte-Carlo and empirical results, respectively. Section \ref{sec:concl} concludes. The appendices contain the proofs of the propositions and some additional empirical results.

\section{Literature review}\label{sec:lit}

%Change-points (CP) are abrupt changes in the generative parameters of a time-series and the their detection is called change-point detection (CPD). Our approach detects CP in the mean of the time series in an online manner allowing for dependent data within regimes and time-varying parameters evolving across regimes. In this section, we provide a brief overview of the CPD literature and we highlight what distinguishes our contributions.

The literature on CPD is extensive, encompassing both theoretical and practical aspects over a long history. Early work in this field can be traced back to Page's contributions \cite{Page} to the quality control literature, where he introduced a method known as the cumulative sum (CUSUM). This method detects change-points by signaling when the cumulative sum of observations exceeds a certain threshold, indicating a shift in the data. Let us notice that the method is devised in an offline setting: any time a new data point is observed, solving the inference problem of change-points requires the method to be applied to the whole dataset again; as such, it is not well suited for real-time applications. 

Since Page's era, the problem of CPD has garnered interest across various research fields, with significant activity in the machine-learning community. This is particularly evident in applications such as signal processing and object tracking; see, e.g., \cite{staudacher,survey2017,truong} to name but a few. Identifying abrupt changes is crucial across many disciplines, including finance, where it is often referred to as regime detection (see, e.g., \cite{Hamilton2010}).

As a result, CPD has been approached in diverse ways to meet the specific needs of different research areas. Despite these varied approaches, they can generally be categorized into three main classes: (i) online vs. offline methods, (ii) methods for univariate vs. multivariate time series, and (iii) parametric vs. non-parametric methods. Here, we focus on two main streams of research: machine learning and econometrics. We also categorize each method within these streams into one of the three principal approaches mentioned above.

Research within the machine learning community has led to the development of both parametric and non-parametric methods for CPD. Parametric methods require specifying the functional form of the data, such as the probability distribution of the data-generating process (DGP). These models learn the parameters of the distributions and identify change-points by detecting shifts in the distribution within a rolling window. For instance, \cite{Kawahara}, \cite{Cho}, and \cite{kucharczyk} propose offline methods and analyze the probability distribution of data before and after a candidate CP by comparing the logarithm of the likelihood ratios of two consecutive time-series intervals. Moreover, \cite{kucharczyk} also study the problem of including a time-varying variance to account for heteroskedastic effects in the data. Another example is the Bayesian Online Change-Point Detection (BOCPD) method (\cite{c:07}), which is the starting point of our methodological innovations. This parametric approach assumes data independence and employs a message-passing algorithm to recursively compute the posterior distribution of the time since the last CP, thus improving upon earlier ideas developed by \cite{r:07}. Based on a Bayesian inference framework, a clear advantage of the proposed approach relies on online learning, that is the updating of the model's parameters any time a new observation is collected, including the update of the probability that a CP has occurred. This makes the method suitable for real-time applications.

On the other hand, non-parametric approaches to CPD do not assume a specific functional form for the DGP. Instead, a CP is typically identified when a test statistic exceeds a certain threshold, signaling a high probability that a CP has occurred. This category includes kernel-based methods, such as those proposed by \cite{Harchaoui2,Harchaoui}, where the authors utilize a Kernel Fisher Discriminant Ratio test statistic, which is based on the kernelized version of the Fisher discriminant analysis, to test for homogeneity between two sliding windows of independent data, or by \cite{ge}, which leverage kernel estimation of CPs to address the problem of causality between time series in a regime-switching context. Even though the kernel-based methods are flexible enough due to their non-parametric nature, a common drawback of these methods is that their performance heavily depends on the choice of the kernel and its parameters, which are being learned by using a training set, thus making these methods suitable for an offline approach. Another line of research within non-parametric approaches involves test statistics utilizing graph-based methods (see, e.g., \cite{FriedmanandRafsky,Rosenbaum,GPAnnStat}). In this case, a graph is constructed for each pair of time series data samples to be compared. The number of edges connecting observations from these samples serves as an indicator of a CP, with fewer edges suggesting a higher likelihood of a change. For example, \cite{Rosenbaum} proposed
minimum distance pairing in which divides the time-series into two equally sized non-overlapping sub-samples in such a way as to minimize the total distances between the sub-samples. This method has the desirable property of being truly
non-parametric and can be applied to multivariate time series. However, it is an offline method that works under the assumption of data independence. Similarly, statistical tests for detecting distribution
changes in multivariate data streams by means
of histograms have also been applied in \cite{QuantTree}. In particular, they detect changes by using a hypothesis test, which assesses whether the data of study is consistent with a
reference histogram learned from a training set, thus making this method to be offline. Recently, \cite{Zou24} employ a deep reinforcement learning approach within the machine learning framework to address robot navigation in changing environments, using an offline method with a finite number of regimes on spatio-temporal data.

In the econometric literature, research focuses predominantly on parametric methods, which are typically devised in an offline framework. However, these methods generally deal with an autoregressive structure for the DGP, rather than assuming independence of observations as in machine learning methods. The seminal work by \cite{Hamilton1988,Hamilton1989} introduced the Markov Switching model and has been widely applied in economic and financial contexts. This autoregressive model features a process mean that switches between a finite number of predefined regimes. The current regime displays a dependence on the previous one, hence the Markov property. It is an autoregressive extension of hidden Markov models (HMM), whose estimation is based on a message-passing algorithm initially introduced in the speech recognition literature (see \cite{Lindgren,Baum}).

Later, motivated by the long-range correlations exhibited by the GARCH model in samples with structural changes (see, e.g., \cite{MikoschStarica2004}), \cite{Cai,HamiltonSusmel} proposed a Markov Switching-Autoregressive Conditional Heteroskedastic model, along with a Markov switching GARCH model by \cite{Gray}. Estimating Markov switching GARCH models using the maximum likelihood method is practically infeasible because the conditional variance depends on the entire past history of the state variables, a problem known as path dependence. To address this computational challenge, simulation-based techniques are employed. \cite{Bauwens} and \cite{EfficientGibbssampling} use Bayesian inference, treating latent state variables as model parameters, thereby allowing the likelihood function to be computed as if the states were known. \cite{He} used the same technique to detect CPs in the GARCH model in real-time focusing on the detection of structural breaks in variance, thereby differentiating it from our method, while \cite{Casarin2024} has generalized it to multivariate time series.

In this paper, we propose a methodology that blends machine learning and econometric literature within an online framework. We build upon the work of \cite{c:07} to devise a general methodology for CPD for the mean of a time series, accounting both for temporal dependence of observations and for time-varying parameters (other than the mean), thus capturing complex dynamic patterns in the data. In fact, we relax two major assumptions of \cite{c:07}, namely i.i.d. random variables as DGP within a regime and constant parameters within regimes. Specifically, we extend BOCPD to accommodate an autoregressive structure of any order within regimes and incorporate the class of Score-Driven models (see \cite{Score-Driven1, Harvey}) to introduce time-varying variance and correlation across regimes. Score-driven models assume autoregressive dynamics for the parameters (e.g., variance and correlation) coupled with an innovation term proportional to the score, i.e. the derivative of the loglikelihood of data, which drives their evolution any time a new data point is observed, in the direction of the maximization of the likelihood. This  class of models is rich in describing dynamic patterns in data, encompassing many well-known models like the GARCH. Moreover, the estimation process is based on Bayesian inference, similar to the approaches by \cite{Bauwens, EfficientGibbssampling}, allowing for fast computation of the likelihood. In conclusion, the method we propose represents a cost-effective solution in computational terms when compared to standard approaches and preserves further the online framework, resulting in a very flexible method suited for real-world applications with general assumptions on the dependence structure, possibly time-varying, of time series with CPs for the mean.

%=================================
\section{Methods}
\label{sec:methods}
In this section, we introduce our methodology of CPD of a univariate time series where regimes have different mean. Comparing to the existing methods, our proposal displays a number of advantages: the proposed probabilistic method detects change-points (i) in an online fashion, (ii) without imposing {\it a priori} a finite number of states for the mean, (iii) accounting for general temporal dependence structure of any memory order $q$ within regimes, and (iv) describing time-varying patterns for the other moments of the distribution like variance and autocorrelation. We start by reviewing the BOCPD model introduced in \cite{c:07}, upon which we build up the novel methodology. In particular, the baseline BOCPD model is generalized as follows. In the first step, we relax the assumption of data independence within each regime by assuming an autoregressive structure for the time series data of any order $q\in\N^*$. We name MBO($q$) this class of models. In a second step, we consider the Markovian specification of this class, namely MBO(1), and promote the parameters controlling for either the autocorrelation or the variance within a regime to be time-varying parameters evolving as the Score-Driven model introduced by \cite{Score-Driven1,Harvey}. We term these new models MBOC (for time-varying autocorrelation) and MBOV (for time-varying variance).

\subsection{Bayesian Online Change-Point Detection}
\label{subsec:BOCPD}
BOCPD, introduced by \cite{c:07}, is a CPD method that exhibits a number of advantages. First, regime detection occurs {\it online} and, as such, BOCPD is suitable for real-time applications. Moreover, the number of regimes is not set {\it a priori}, but CP defines a change for the value of the parameters of the model, which adapts to the regime for the best description of the time series data. Finally, BOCPD is devised for any distribution of the exponential family, thus achieving high flexibility in data description. The method works as follows. 
\subsubsection{General framework}
Let $\{\cdots,x_{-2},x_{-1},x_0,x_1,\cdots\}\in\mathbb{R}^{\mathbb{Z}}$ be a time series and consider the observed sample $x_{1:T} = \{x_1,...,x_T\}$. The model introduced by \cite{c:07} assumes that $\{x_i\}_{i=1,\ldots,T}$ are i.i.d. random variables in each regime. Data are assumed to follow a product partition model (PPM), see e.g. \\\cite{r:92}, meaning that (i) data can be partitioned into regimes, (ii) sub-samples of data are stationary within each regime, and (iii) regimes are independent. A distribution $P$ in the exponential family generates sub-samples of data and two different and non-overlapping sub-samples $R_1$ and $R_2$ are generated by the same distribution $P$ with two different parameters $\theta_{R_1}$ and $\theta_{R_2}$. More precisely, the parameters $\theta_{R}$ defining each regime $R$ are i.i.d. random variables drawn from some probability distribution $p(\theta_R|\eta_0)$ where $\eta_0$ are the prior hyperparameters. For the sake of analytical tractability, authors used conjugate prior distributions, namely distributions that have the same functional form with the posterior distributions of the parameters. It has been proven \cite{r:79} that any probability distribution of the exponential family admits a conjugate prior. Leveraging the conjugacy property, the prior is chosen in such a way the posterior distribution keeps the same functional form as the prior. As such, the prior distribution belongs to the exponential family as well.

Regimes and CPs defining when a regime has started are not directly observable, but they represent hidden variables that need to be inferred from data. To this end, the goal is to estimate the elapsed time since the last CP, a quantity named the {\it run length} of a regime. It is defined as follows.
\begin{definition}
    The run length $r_t$ is a non-negative discrete variable defined as:
    \begin{equation}
    r_t = 
    \begin{cases}        0,\quad\quad\quad\quad\text{if a CP occurs at time $t$,}\\
        r_{t-1}+1,\quad\text{otherwise}.
    \end{cases}
\end{equation}
\end{definition}

In the general context of Bayesian inference, the aim is to find first a closed-form expression for the posterior distributions of parameters, namely $p(r_t\vert x_{1:t})$ for the run length and $p(\theta_R\vert x_{1:t})$ for the parameters of the exponential family's distribution, then update such posterior distributions any time a new data point is observed.\footnote{For notational simplicity, we consider here that the last CP has been at time $t=0$ and we are observing data points in the new regime before the occurrence of a new CP.}

BOCPD models the arrival of a CP as a Bernoulli process
with hazard rate $1/h$, that is the probability of a run length $r_t$ conditional to $r_{t-1}$ (also named the {\it hazard}) is
\begin{equation}\label{eq: hazard}
    p(r_t|r_{t-1}) = \begin{cases}
        1/h,\quad\quad\quad\text{if $r_t=0$,}\\
        1-1/h,\quad\text{if $r_t = r_{t-1}+1$,}\\
        0,\quad\quad\quad\quad\text{otherwise.}
    \end{cases}
\end{equation}

Since $r_t$ is a hidden variable, the inference method is based on the maximization of the run length posterior $p(r_t|x_{1:t})$, namely the probability of some value $r_t$ given the observations $x_{1:t}$ from the last CP up to the current time step $t$. From the Bayes' rule, the run length posterior can be stated as
\begin{equation}\label{eq:RL posterior}
    p(r_t|x_{1:t}) = \frac{p(r_t,x_{1:t})}{p(x_{1:t})}.
\end{equation}
The quantity $p(x_{1:t})$  is named {\it evidence} (of data) and is computed as
\begin{equation}\label{eq:evidence}
    p(x_{1:t}) = \sum_{r_t}p(r_t,x_{1:t}).
\end{equation}
The joint distribution $p(r_t,x_{1:t})$ of both the run length and the observed data can be further stated in a recursive form by using marginalization and Bayes' rule as
\begin{align}\label{eq:joint distr}
    p(r_t,x_{1:t}) & = 
        \sum_{r_{t-1}}\underbrace{p(x_t|x_{t-1}^{(r_{t-1})})}_\text{UPM}\underbrace{p(r_t|r_{t-1})}_\text{Hazard}\underbrace{p(r_{t-1},x_{1:t-1})}_\text{Message}
\end{align}
where $p(x_t|x_{t-1}^{(r_{t-1})})$ is named the  {\it Underlying Predictive Model} (UPM), and the joint distribution $p(r_{t-1},x_{1:t-1})$ up to time $t-1$ is the {\it Message}. In particular, the UPM is the predictive posterior distribution given the current run length. Because of the assumption on PPM, such a distribution depends only on the last $r_{t-1}$ observations and can be stated in a more compact form as
\begin{align}
    p(x_t|r_{t-1},x_{1:t-1}) & = p(x_t|x_{t-1}^{(r_{t-1})})
\end{align}
where
\begin{equation}
    x_{t-1}^{(r_{t-1})} = 
        x_{t-r_{t-1}:t-1}
    \quad\text{and}\quad x_{t:t-1} = \emptyset .
\end{equation}

Using the exponential family's conjugacy property when data are i.i.d., one can obtain closed-form solutions for the UPM term; see \cite{r:79, k:07}. In particular, the aim is to infer $\theta_R$ by studying the posterior distribution,
\begin{equation}
    p(\theta_R|x_{t-1}^{(r_{t-1})})\propto p(x_{t-1}^{(r_{t-1})}|\theta_R)p(\theta_R|\eta_0).
\end{equation}
By choosing a conjugate prior $p(\theta_R|\eta_0)$, the posterior distribution takes the following explicit form,
\begin{equation}
    p(\theta_R|x_{t-1}^{(r_{t-1})})\propto p(\theta_R|\eta_{r_{t-1}}),
\end{equation}
where $\eta_{r_{t-1}}$ are the (posterior) hyperparameters whose value has been updated using the observations in the last run length.

Then, the UPM term becomes a function of the hyperparameters as follows,
\begin{align}
    p(x_t|x_{t-1}^{(r_{t-1})}) & = \int p(x_t|x_{t-1}^{(r_{t-1})},\theta_R)p(\theta_R|x_{t-1}^{(r_{t-1})})d\theta_R \nonumber \\
    & = \int p(x_t|\theta_R)p(\theta_R|x_{t-1}^{(r_{t-1})})d\theta_R \nonumber \\
    & \propto \int p(x_t|\theta_R)p(\theta_R|\eta_{r_{t-1}})d\theta_R \nonumber \\
    & = p(x_t|\eta_{r_{t-1}}).
\end{align}
Notice that the second equality follows from the fact that $x_t$ is conditionally independent of the past data point, given the parameters, hence $p(x_t|x_{t-1}^{(r_{t-1})},\theta_R) = p(x_t|\theta_R)$.

Finally, an assumption simplifying the computation is the conditional dependence of the CP prior, namely $r_t$ is conditionally dependent on $r_{t-1}$ only. As such, the posterior probability for $r_t$ varies depending on $r_{t-1}$ or, in other words, it depends largely on the last most likely CP, whose probability is in Eq. (\ref{eq:RL posterior}). Since the value of the run length is not deterministic but probabilistic, the most likely value of the run length is used to define a regime, as stated below.

\begin{definition}\label{def:regime}
Let $x_{1:T}$ be a time series and $t,s\in\N\cap[1,T]$ times with $t<s$ such that 
\begin{equation}
    \argmax_{i\in\{0,1,...,t\}}p(r_t=i|x_{1:t}) = \argmax_{i\in\{0,1,...,s\}}p(r_s=i|x_{1:s}) = 0
\end{equation}
 and for any $u\in\N\cap(t,s)$, $\argmax_{i\in\{0,1,...,u\}}p(r_u=i|x_{1:u}) \neq 0$. Then the subset $x_{t:s-1}$ of the time series $x_{1:T}$ is defined as a regime.
\end{definition}

\subsubsection{Gaussian i.i.d. case}\label{subsec:normal_case} To proceed further, one must specify one particular probability distribution belonging to the exponential family and which parameter is different in each regime. \cite{c:07} considered the case of data as i.i.d. Gaussian random variables and regimes defined depending on the value of the mean of the normal distribution after a CP, assuming the value of the variance is known and constant over all regimes.\footnote{In the practical implementation of the BOCPD, the variance is computed over some training set before applying the method. Below, we show how to relax such an assumption by proposing an observation-driven description of parameters like variance and auto-correlation, which are thus promoted to time-varying parameters.}

Let a time series be the sequential realization of i.i.d. random variables from a normal distribution with unknown mean $\theta_{R}$ and known variance $\sigma^2$ within regime $R$,
\begin{equation}\label{eq4}
    x_i\sim\mathcal{N}(\theta_{R},\sigma^2).
\end{equation}
By using the conjugate prior for the mean, it is
\begin{equation}
    p(\theta_R|\eta_0) = \mathcal{N}(\theta_R|\underbrace{\mu_0,\sigma_0^2}_{\eta_0})
\end{equation}
where $\eta_0=\{\mu_0,\sigma_0\}$ are the hyperparameters of the prior distribution of $\theta_R$. \cite{c:07} obtained the closed form of the UPM term as
\begin{align}\label{eq13}
    p(x_{t}|x_{t-1}^{(r_{t-1})}) & = \mathcal{N}(x_t|\underbrace{\mu_{r_{t-1}},\sigma^2+\sigma_{r_{t-1}}^2}_{\eta_{r_{t-1}}}),
\end{align}
where $\eta_{r_{t-1}}=\{\mu_{r_{t-1}},\sigma^2+\sigma_{r_{t-1}}^2\}$ are the hyperparameters of the posterior distribution of $\theta_R$, which has been updated using observations $x_{t-1}^{(r_{t-1})}$ in the last regime. In particular, the posterior hyperparameters are 
\begin{equation}\label{eq: post params BOCPD}
    \mu_{r_{t-1}} = \frac{\frac{\sum_{i=t-r_{t-1}}^{t-1}x_i}{\sigma^2}+\frac{\mu_0}{\sigma_0^2}}{\frac{r_{t-1}}{\sigma^2}+\frac{1}{\sigma_0^2}}\quad\text{and}\quad \sigma_{r_{t-1}}^2 = \Big(\frac{r_{t-1}}{\sigma^2}+\frac{1}{\sigma_0^2}\Big)^{-1}\quad\text{for}\quad r_{t-1}\in\{1,...,t-1\}.
\end{equation}
%In Appendix \ref{app: BOCPD Derivations}, we review the proof of the above expressions which were derived in \cite{c:07}. {\bf Is this Appendix necessary? I don't see why to repeat a proof shown in another paper.}

Finally, the hazard rate is a hyperparameter that is tuned  by using a small sample of the time-series.

\subsubsection{BOCPD model}\label{subsec:BOCPD_algorithm}
The BOCPD model (see Algorithm \ref{alg: BOCPD}) works as follows. At time $t=0$, one initializes the prior values $\mu_0, \sigma_0^2$ and the known variance $\sigma^2$. At generic time $t>0$, a new data point $x_t$ becomes available, and the UPM in Eq.~(\ref{eq13}) is computed for any possible $\mu_{r_{t-1}}$ and $\sigma^2_{r_{t-1}}$, as a function of the run length $r_{t-1}$ that takes value from $0$ to $t-1$. Then, the joint distribution over both the run length and the observed data point, see Eq. (\ref{eq:joint distr}), is computed for all the possible values of the run length. \cite{c:07} obtained:
\begin{enumerate}
    \item The growth probabilities,
    \begin{equation}
       p(r_t=l,x_{1:t}),\quad\text{for } l = 1,...,t;
    \end{equation}
    \item The CP probability
    \begin{equation}
        p(r_t=0,x_{1:t}).
    \end{equation}
\end{enumerate}
After computing the evidence in Eq. (\ref{eq:evidence}), the run length posterior is obtained via Eq.~(\ref{eq:RL posterior}). Finally, $\mu_{r_t}$ and $\sigma^2_{r_t}$ are updated as in Eq. (\ref{eq: post params BOCPD}) in view of the next step at time $t+1$.\footnote{A Python implementation of the method is available online at \url{https://gitlab.com/YvTsak/ScoreDrivenBOCPD}.}

\begin{algorithm}[tb]
\caption{BOCPD}
\label{alg: BOCPD}
\textbf{Input}: $\mu_0,\sigma_0^2, p(r_0=0) = 1$\\
\textbf{Output}: $p(r_t|x_{1:t}), \hat{\mu}_t$
\begin{algorithmic}[1] %[1] enables line numbers
\For{$t = 1,...$} 
\State Observe $x_t\sim\mathcal{N}(\theta_{R},\sigma^2)$
\State Compute $\hat{\mu}_t$
\State Compute $p(r_t|x_{1:t})$
\State Update $\mu_t,\sigma_t^2$
\EndFor
\end{algorithmic}
\end{algorithm}

%=================================
\subsection{MBO($q$) model}%: temporal dependence in time series}
Here, we extend the baseline framework in \cite{c:07} to the case of temporal dependence in time series data, thus relaxing the assumption of i.i.d. random variables within a regime. We introduce the MBO($q$) class of models that extends the BOCPD model to the case of dependent data of memory order $q\in\N^*$ within regimes. The key observation is that the conjugacy property still holds when the data are auto-correlated since the conditional distribution of any member in the exponential family is still in the family, see \cite{r:08}. 
% Let $\Sigma_\infty = (\gamma_{s-t})_{s,t=-\infty}^\infty$ be the covariance matrix of the time-series of study. Throughout the paper we will write $\Sigma_j = (\gamma_{s-t})_{1\leq s,t\leq j}$ to denote the finite covariance matrix of dimension $j\times j$ for $j\in \N^*$. 

Let $x_i$ and $x_{i-j}$ be two data points belonging in the same regime $R$. We build the MBO($q$) class of models upon two main assumptions:
\begin{itemize}
    \item [(A1)] The process is covariance stationary in each regime and the covariance structure is the same in all regimes.%For every regime $R$, the covariance at lag $j$ for $j\in\{0,\cdots,q\}$ is $\gamma_j^R$.
\end{itemize}
Specifically, if $\gamma_{ij}^R = \cov(x_i,x_{i-j}|x_i,x_{i-j}\in R)$ is the covariance at lag $j$ at time $i$, we assume that $\gamma_{ij}^R=\gamma_j$. We define the $j\times j$ matrix $\Sigma_j=(\gamma_{s-t})_{1\leq s,t\leq j}$. 
\begin{itemize}
    \item [(A2)] Within a regime $R$, the DGP\footnote{Within the exponential family, we consider here the Gaussian distribution as DGP. In general, any other member from the exponential family can be considered.} is as follows:
    \begin{itemize}
        \item For $i\in\N^*$
        \begin{equation}
             x_i \sim\mathcal{N}(\theta_R,\gamma_0).
        \end{equation}
        \item For $i>1$ and $j^* = \min\{q,i-1\}$
                \begin{align}\label{eq: A2 conditional distribution}
x_i|x_{i-1}^{(j^*)}\sim\mathcal{N}\left(\theta_R+\begin{bmatrix}
        \gamma_1\\
        \vdots \\
        \gamma_{j^*}
    \end{bmatrix}^\intercal\Sigma_{j^*}^{-1}\begin{bmatrix}
        x_{i-1}-\theta_R\\
        \vdots\\
        x_{i-j^*}-\theta_R
    \end{bmatrix},\underbrace{\gamma_0-\begin{bmatrix}
        \gamma_1\\
        \vdots \\
        \gamma_{j^*}
    \end{bmatrix}^\intercal\Sigma_{j^*}^{-1}\begin{bmatrix}
        \gamma_1\\
        \vdots\\
        \gamma_{j^*}
    \end{bmatrix}}_{v_{j^*}}\right)
\end{align}
    \end{itemize}
\end{itemize}
i.e., similarly to section \ref{subsec:normal_case}, we focus on the case of Gaussian probability distributions. Note that $x_{i-1}^{(1)} = x_{i-1:i-1} = x_{i-1}$. In the rest, we will use the notation $v_{j^*}$\footnote{Note that $v_{j^*}\in\R$.} for the conditional variance in Eq.~(\ref{eq: A2 conditional distribution}) with $v_0 = \gamma_0$. 
Assumption (A2) is satisfied by any linear autoregressive model, for example, assume that the data within regime $R$ is a realization of an AR($1$) model, then it is:
\begin{equation}
    x_i = \mu_0+\phi_1 x_{i-1}+\epsilon_i,\quad\epsilon_i\overset{\mathrm{iid}}{\sim}\mathcal{N}(0,\sigma^2).
\end{equation}
In this case, we observe that assumption (A2) is satisfied because
\begin{align}
    x_i&\sim\mathcal{N}\left(\frac{\mu_0}{1-\phi_1},\frac{\sigma^2}{1-\phi_1^2}\right),\\\label{eq: conditional distribution ar(1)}
    x_i|x_{i-1}&\sim\mathcal{N}(\mu_0+\phi_1 x_{i-1},\sigma^2).
\end{align}
By using the Yule-Walker equations associated with the process above (see, e.g., \cite{Hamilton}) we get
\begin{equation}
     \gamma_0 = \frac{\sigma^2}{1-\phi_1^2}\quad\text{and}\quad
    \gamma_1 = \frac{\phi_1\sigma^2}{1-\phi_1^2}.
\end{equation}
In particular, computing the conditional variance $v_1$ from Eq. (\ref{eq: A2 conditional distribution}) we get the conditional variance in Eq.~(\ref{eq: conditional distribution ar(1)}) as
\begin{align}
    \gamma_0-\frac{\gamma_1^2}{\gamma_0} & = \sigma^2.
\end{align}
Moreover, using $\theta_R = \frac{\mu_0}{1-\phi_1}$ and Eq.~(\ref{eq: conditional distribution ar(1)}), the conditional mean in Eq.~(\ref{eq: A2 conditional distribution}) results as
\begin{equation}
    \theta_R+\frac{\gamma_1}{\gamma_0}(x_{i-1}-\theta_R) = \mu_0+\phi_1 x_{i-1}.
\end{equation}
The next proposition shows how to obtain the UPM term and the posterior hyperparameters for the MBO($q$) model.
\begin{proposition}\label{prop: MBO(q)}
Let us assume (A1) for the sample time series $x_{1:T}$. Then, for any regime $R$ determining a sub-sample realization $x_t^{(r_t)}$ satisfying assumption (A2), by using a conjugate prior for the mean, namely $p(\theta_R)\sim\mathcal{N}(\mu_0,\sigma_0^2)$, the UPM term exists in closed form, and it is
\begin{equation}
   p(x_{t+1}|x_{t}^{(r_t)})
     = \mathcal{N}(\pmb{1}^{q+1}c_{q+1}\mu_{r_t}+\begin{bmatrix}
        \gamma_1\\
        \vdots\\
        \gamma_q
    \end{bmatrix}^\intercal\Sigma_q^{-1}\pmb{x}_{t-1}^{(q)},v_{q}+\pmb{1}^{q+1}c_{q+1}\sigma_{r_t}^2(\pmb{1}^{q+1}c_{q+1})^\intercal)
\end{equation}
where $\pmb{1}^{q+1}$ is the unit vector of dimension $q+1$ and $\pmb{x}_{t-1}^{(q)} = \begin{bmatrix}
    x_{t-1}\\
    \vdots\\
    x_{t-q}
\end{bmatrix}$. Then the posterior hyperparameters are
\begin{equation}
    \sigma_{r_t}^2  = \frac{1}{a_{r_t}}\quad\text{and}\quad
    \mu_{r_t}  = \frac{b_{r_t}}{a_{r_t}}
\end{equation}
with
\begin{equation}
    a_{r_t} = \sum_{1\leq i,j\leq s} a^{r_t}_{ij}\quad\text{and}\quad b_{r_t} = \sum_{1\leq j\leq s} b^{r_t}_{j},\quad\text{for}\quad s=\frac{(q+1)(q+2)+1}{2}
\end{equation}
where the terms in the summation are from the matrices
\begin{align}
    A_{r_t} & = \begin{pmatrix}
        \frac{(r_t-q)C_{q+1}}{v_{q}} & 0 & 0 & \cdots & 0 & 0 & 0\\
        0 & \frac{C_{q}}{v_{q-1}} & 0 & \cdots & 0 & 0 & 0\\
        0 & 0 & \frac{C_{q-1}}{v_{q-2}} & 0 & \cdots & 0 & 0\\
        0 & 0 & 0 & \frac{C_{q-2}}{v_{q-3}} & \cdots & 0 & 0\\
        \vdots & \vdots & \vdots & \cdots & \vdots & \vdots & \vdots\\
        0 & 0 & 0 & \cdots & 0 & \frac{C_1}{v_{0}} & 0\\
        0 & 0 & 0 & \cdots & 0 & 0 &\frac{1}{\sigma_0^2}
\end{pmatrix} \quad\text{where}\quad A_{r_t} = (a^{r_t}_{ij})_{1\leq i,j\leq s}\in\R^{s\times s}
\end{align}
and
\begin{align}
    b_{r_t}^T = \begin{pmatrix}
        \sum_{i=q+1+t+r_t}^t x_i\\
        \vdots\\
        \sum_{i=q+1+t+r_t}^t x_{i-q}\\
        \pmb{x}_{q+t-r_t}^{(q)}\\
        \vdots\\
        \pmb{x}_{1+t-r_t}^{(1)}\\
        1
    \end{pmatrix}^T \begin{pmatrix}
         \frac{C_{q+1}}{v_{q}} & 0 & 0 & \cdots & 0 & 0 & 0\\
         0 & \frac{C_{q}}{v_{q-1}} & 0 & \cdots & 0 & 0 & 0\\
        0 & 0 & \frac{C_{q-1}}{v_{q-2}} & 0 & \cdots & 0 & 0\\
        0 & 0 & 0 & \frac{C_{q-2}}{v_{q-3}} & \cdots & 0 & 0\\
        \vdots & \vdots & \vdots & \cdots & \vdots & \vdots & \vdots\\
        0 & 0 & 0 & \cdots & 0 & \frac{C_1}{v_{0}} & 0\\
        0 & 0 & 0 & \cdots & 0 & 0 & \frac{\mu_0}{\sigma_0^2}
    \end{pmatrix}\quad\text{where}\quad b_{r_t}^T = (b^{r_t}_{j})_{1\leq j\leq s}\in\R^{1\times s}
\end{align}
with
\begin{equation}\label{eq:Cjplus1}
C_{j+1} = c_{j+1}c_{j+1}^\intercal\:\: \mbox{where}\:\: c_{j+1} = \begin{bmatrix}
    1\\
    -\Sigma_{j}^{-1}\begin{bmatrix}
        \gamma_1\\
        \vdots\\
        \gamma_j
    \end{bmatrix}
\end{bmatrix}\:\: \mbox{for}\:\: j\in\{1,\cdots,q\}.
\end{equation}
\end{proposition}
The proof of the proposition is in Appendix \ref{app: MBO(q) Derivations}. Notice that by setting $\gamma_j=0$ for every $j\in\{1,\cdots,q\}$ we recover the UPM and the posterior hyperparameters of the independent case, see Eq. (\ref{eq13}-\ref{eq: post params BOCPD}).
%=================================
\subsubsection{MBO(1) model}\label{subsec: MBO(1)}%: Markovian data}
Let us focus here on the case $q=1$, the Markovian extension MBO(1) of the BOCPD model. We assume that the DGP for the time series satisfies assumptions (A1) and (A2). In other words, the data $x_t^{(r_t)}$ within a regime $R$ are described as
\begin{align}
    x_{t+1-r_t} & \sim \mathcal{N}\label{eq:Markov dependence 1}(\theta_{R},\gamma_0),\\\label{eq:Markov dependence 2}
    x_i|x_{i-1} & \sim \mathcal{N}\left(\theta_{R}+\frac{\gamma_1}{\gamma_0}(x_{i-1}-\theta_{R}),\gamma_0\left(1-
    \left(\frac{\gamma_1}{\gamma_0}\right)^2\right)\right),
\end{align}
for any $i\in(t+1-r_t,t]$ and for any $r_t$.

As in the baseline model, the unconditional distribution is normal within each regime, with unknown mean $\theta_{R}$ and known variance $\gamma_0$. Moreover the conditional distribution is normal with constant correlation $\gamma_1/\gamma_0$. The corollary below provides the explicit solution for the UPM term and the posterior hyperparameters in the Gaussian auto-correlated case of lag 1.

\begin{corollary}\label{prop: MBO}
    Assume that in a regime $R$ the time series $x_t^{(r_t)}$ is described by Eq. (\ref{eq:Markov dependence 1})-(\ref{eq:Markov dependence 2}). Then, by using a conjugate prior for the mean, namely $p(\theta_R)\sim\mathcal{N}(\mu_0,\sigma_0^2)$, the closed form expression of the UPM term is
    \begin{align}
    p(x_{t+1}|x_t^{(r_t)})
    & = \mathcal{N}\big(\mu_{r_t}+\frac{\gamma_1}{\gamma_0}(x_t-\mu_{r_t}),\gamma_0(1-(\frac{\gamma_1}{\gamma_0})^2)+\sigma_{r_t}^2(1-\frac{\gamma_1}{\gamma_0})^2\big),
\end{align}
where the posterior hyperparameters are updated as
\begin{align}\label{eq: post params MBO1}
    \mu_{r_t} = \frac{b_{r_t}+\frac{\mu_0}{\sigma_0^2}}{a_{r_t}+\frac{1}{\sigma_0^2}}\quad\text{and} \quad\sigma_{r_t}^2 = \Big(a_{r_t}+\frac{1}{\sigma_0^2}\Big)^{-1}\quad\text{for}\quad r_t\in\{1,...,t\}
\end{align}
with
\begin{align}
    a_{r_t} & = \frac{1}{\gamma_0}+\frac{(r_t-1)(1-\frac{\gamma_1}{\gamma_0})^2}{\gamma_0(1-(\frac{\gamma_1}{\gamma_0})^2)}\\
    b_{r_t} & = \begin{cases}\frac{x_{t}}{\gamma_0},\quad\text{for}\quad r_t=1\\[10pt] 
    \frac{x_{t-1}}{\gamma_0}+\frac{(1-\frac{\gamma_1}{\gamma_0})(x_t-\frac{\gamma_1}{\gamma_0} x_{t-1})}{\gamma_0(1-(\frac{\gamma_1}{\gamma_0})^2)},\quad\text{for}\quad r_t=2\\[10pt]
        \frac{x_{t+1-r_t}}{\gamma_0}+\frac{(1-\frac{\gamma_1}{\gamma_0})^2\sum_{i=t+2-r_t}^{t-1}x_i+(1-\frac{\gamma_1}{\gamma_0})(x_t-\frac{\gamma_1}{\gamma_0} x_{t+1-r_t})}{\gamma_0(1-(\frac{\gamma_1}{\gamma_0})^2)},\quad\text{for}\quad r_t\in\{3,...,t\}.
    \end{cases}.
\end{align}
\end{corollary}
The corollary is deduced by setting $q=1$ in proposition \ref{prop: MBO(q)}. 

Let us highlight now the asymptotic behavior of the posterior hyperparameters in the case of a regime of infinite run length. The general question is about the convergence of the hyperparameters $\mu_t$ and $\sigma_t^2$ to some asymptotic value when we collect more and more data points within the same regime $R$. This has been studied for the Gaussian independent case by \cite{Bishop}. Here, we show the extension to Gaussian auto-correlated data of lag 1.

Let be $r_t\equiv t$, then the posterior hyperparameters in Eq. (\ref{eq: post params MBO1}) are
\begin{equation}
    \mu_t = \frac{b_t+\frac{\mu_0}{\sigma_0^2}}{a_t+\frac{1}{\sigma_0^2}}\quad\text{and}\quad\sigma_t^2 = \Big(a_t+\frac{1}{\sigma_0^2}\Big)^{-1}.
\end{equation}
The following proposition states that the distribution of the unknown mean converges asymptotically to the maximum likelihood solution of the corresponding problem in the frequentist approach.
\begin{proposition}\label{prop1}
    Let be $r_t\equiv t$ and let us assume a Markovian structure within the regime. Then the maximum likelihood solution for the unknown mean $\theta_R$ is 
    \begin{equation}
        \theta_{ML} = \frac{x_1+\frac{1-\frac{\gamma_1}{\gamma_0}}{1-(\frac{\gamma_1}{\gamma_0})^2}[(1-\frac{\gamma_1}{\gamma_0})\sum_{i=2}^{t-1}x_i+x_t-\frac{\gamma_1}{\gamma_0} x_1]}{1+\frac{(1-\frac{\gamma_1}{\gamma_0})^2(t-1)}{1-(\frac{\gamma_1}{\gamma_0})^2}}
    \end{equation}
    and the mean value $\mu_t$ of the posterior mean is a weighted average between $\theta_{ML}$ and the prior mean value $\mu_0$.
     Moreover, as the number of data points increases, the mean value of the posterior for the mean hyperparameter converges to the maximum likelihood solution, and the limiting variance of the posterior goes to zero, namely
    \begin{equation}
        \lim_{t\rightarrow\infty}\mu_t = \theta_{ML}\quad\text{and}\quad\lim_{t\rightarrow\infty}\sigma_t^2 = 0.
    \end{equation}
\end{proposition}
The proof is in Appendix \ref{app: Connection with Maximum Likelihood}.
%=================================
\subsection{Score-Driven MBO(1)}\label{subsec:Score-Driven BOCPD}

The proposed MBO(q) detects change-points for the mean, properly accounting for temporal dependence in time series, i.e.  a Markovian autoregressive structure for the MBO(1) specification. As such, the parameter associated with the mean is piecewise constant, identifying a new regime at each change-point. However, the other parameters are generally assumed to be constant within regimes. Such an assumption might be unrealistic in many empirical cases. For example, heteroscedasticity, i.e. time-varying variance, is ubiquitous in financial time series \cite{engle}.

Here, we generalize the MBO(1) framework by introducing time-varying parameters for either the variance or the autocorrelation based on {\it observation-driven models}, see \cite{cox}. This allows us to capture dynamic patterns in time series between different regimes and potentially within each regime. Following \cite{cox}, the class of {\it observation-driven models} describes parameters whose current values are deterministic functions of lagged dependent variables but, unconditionally, are random variables because they are functions of the random realization of the time series process. In this setting, parameters evolve randomly over time but are perfectly predictable one-step-ahead given past information.

We name the new model as {\it Score-Driven MBO(1)}, which is an extension of the BOCPD to the case of Markovian dependence of data (MBO(1) model) and with time-varying parameters (for either the variance or the auto-correlation) within the regime.

\subsubsection{MBOC  model}\label{subsec: MBOC model}

Among the general class of observation-driven models, we use here Score-driven models introduced in \\\cite{Score-Driven1} and \cite{Harvey} to describe the parameters as time-varying. Score-driven models assume that the dynamics of the time-varying parameter(s) is autoregressive with an innovation term that depends on the score. The score is the derivative of the log-likelihood of the data with respect to the parameter(s). The score is then re-scaled by a power of the inverse of the Fisher information matrix (see below), which modulates the importance of the innovation based on the concavity of the log-likelihood. The main idea is that the re-scaled score adjusts the parameter(s) to maximize the likelihood of the observed data. Notably, many standard models in financial econometrics, such as the GARCH, ACD, MEM, etc., are special cases of score-driven models (for more details, see \url{www.gasmodel.com}).

In this section, we extend here the auto-correlated case MBO(1) by making the correlation coefficient $\rho$ (i.e. $\frac{\gamma_1}{\gamma_0}$ according to the notation of subsection \ref{subsec: MBO(1)}) to a time-varying parameter $\rho_t$ described by the Score-Driven version of the AR(1) process, as discussed by \cite{Score-Driven_AR}. We assume the variance of data $\sigma^2$ (i.e. $\gamma_0$ according to the notation of subsection \ref{subsec: MBO(1)}), remains constant and name this extension MBOC. Additionally, we then introduce an online method to estimate both the time-varying parameter $\rho_t$ and the regime characteristics, specifically the mean $\theta_R$ characterizing the regime and the run length $r_t$.

% \begin{algorithm}[tb]
% \caption{MBOC original - likelihood}
% \label{alg: MBOC original - likelihood}
% \textbf{Input}: $\mu_0,\sigma_0^2,d,\vec\lambda_0, \rho_{1,d}, \sigma_i^2, p(r_0=0) = 1, \eta$\\
% \textbf{Output}: $p(r_t|x_{1:t}), \hat{\mu}_t$
% \begin{algorithmic}[1] %[1] enables line numbers
% \For{$t = 1,...$} 
% \State Observe $x_t\sim\mathcal{N}(\theta_{R},\sigma^2)$
% \State Compute $\hat{\mu}_t$
% \State Find $\text{argmax}_{i\in\{0,1,...,t\}}p(r_t=i|x_{1:t})$
% \If{$t>1$ \textbf{\&} $i>\eta$}
% \State Compute $\ln\mathcal{L}_t = \sum_{j=t-i+1}^t(-\frac{1}{2}\ln(2\pi)-\frac{1}{2}\ln\sigma^2-\frac{1}{2\sigma^2}(x_j-\frac{\sum_t x_t^{(i)}}{i}-\rho_{t,d}(x_{j-1}-\frac{\sum_t x_t^{(i)}}{i}))^2)$
% \State Find $\vec\lambda_t=\argmin_{\vec\lambda}\{-\ln\mathcal{L}_t\}$
% \State Filter $\rho_{t,d}$
% \EndIf
% \State Compute $p(r_t|x_{1:t})$\Comment{The correlation of the previous step $\rho_{t-1,d}$ is used here}
% \State Update $\mu_t,\sigma_t^2$
% \EndFor
% \end{algorithmic}
% \end{algorithm}

\begin{algorithm}[tb]
\caption{MBOC}
\label{alg: MBOC new}
\textbf{Input}: $\mu_0,\sigma_0^2,d,\vec\lambda_0, \rho_{1,d}, \sigma_i^2, p(r_0=0) = 1, \eta$\\
\textbf{Output}: $p(r_t|x_{1:t}), \hat{\mu}_t$
\begin{algorithmic}[1] %[1] enables line numbers
\For{$t = 1,...$} 
\State Observe $x_t\sim\mathcal{N}(\theta_{R},\sigma^2)$
\State Compute $\hat{\mu}_t$
\State Find $\text{argmax}_{i\in\{0,1,...,t\}}p(r_t=i|x_{1:t})$
\State Remove the mean from $x_{t}^{(i)}$ and get the data $y_t^{(i)}$
\Comment{$y_t^{(i)}$ is the de-meaned data of $x_t^{(i)}$}
% \State Set $y_t = \Tilde{x}_t$ with \Comment{$\Tilde{x}_t$ is the de-meaned data computed in the previous step}
% \State Compute $y_t^{(i)} = x_t^{(i)}-\frac{\sum_t x_t^{(i)}}{i} = \{x_{t-i+1}-\frac{\sum_t x_t^{(i)}}{i},\cdots,x_t-\frac{\sum_t x_t^{(i)}}{i}\}$
\If{$t>\eta$}
\State Infer $\vec\lambda_t$ with GAS using $y_t^{(t)}$
\State Filter $\rho_{t,d}$
\EndIf
\State Compute $p(r_t|x_{1:t})$\Comment{The correlation of the previous step $\rho_{t-1,d}$ is used here}
\State Update $\mu_t,\sigma_t^2$
\EndFor
\end{algorithmic}
\end{algorithm}
More specifically, within a regime $R$, the data-generating process is assumed to be
\begin{align}\label{eq21}
    x_t = \rho_{t,d}(x_{t-1}-\theta_{R})+\theta_{R}+u_t,\quad u_t\sim\mathcal{N}(0,\sigma^2),
\end{align}
where $\theta_{R}$ and $\sigma^2$ are unknown and must be estimated.\footnote{Let us point out that including a Score-Driven model solves implicitly the issue about the unknown (constant) variance $\sigma^2$ in the MBO(1) model. In fact, $\sigma^2$ should be estimated in a pre-training set in order to implement the MBO(1) model. The MBOC generalization allows us to find regimes in the mean, and, at the same time, the parameter $\sigma^2$ is estimated online via maximum likelihood methods.} According to the Score-Driven AR(1) process, the time-varying correlation $\rho_{t,d}$ is described by the recursive relation\footnote{This specification does not guarantee that $|\rho_t|\le 1$, thus sometimes one uses a link function (e.g. an inverse logistic) which maps $[-1,1]$ in ${\mathbb R}$, see \cite{Score-Driven_AR}. In our empirical analysis, we observe that the filtered $|\rho_t|$ is larger than $1$ in less than one per thousand observations. To overcome the issue, we set a threshold $|\rho_t|\leq 1$.}
\begin{equation}\label{eq:scoredrivenrho}
    \rho_{t,d} = \omega+\alpha s_{t-1,d} + \beta\rho_{t-1,d},
\end{equation}
where $s_t$ is the re-scaled score defined as 
\begin{align}\label{eq: I}
    s_{t,d} & = \mathcal{I}_{t|t-1}^{-d}\cdot \nabla_t,\quad d\in[0,1]\\ \label{eq: II}
    \nabla_t & = \frac{\partial \log p_u(u_t)}{\partial \rho_t},\\
    \mathcal{I}_{t|t-1} & = -\E_{t|t-1}[\nabla_t^T\nabla_t],
\end{align}
and $u_t = x_t-\theta_{R}-\rho_t (x_{t-1}-\theta_{R})$ is the prediction error associated with the observation $x_t$.
In the analysis below, for the computation of the score, we set $d=0$ or $d=\frac{1}{2}$.
\begin{itemize}
    \item When $d=0$ there is not a re-scaling factor for the score, that is
    \begin{align}
    s_{t,0} = \nabla_t & = \frac{u_t}{\sigma^2}(x_{t-1}-\theta_{R}).
\end{align}
\item When $d = \frac{1}{2}$ the score is
\begin{align}
    s_{t,1/2} & = (-\E_{t|t-1}[\nabla_t^T\nabla_t])^{-1/2}\nabla_t=\Big(-\E_{t|t-1}\Big[\frac{\partial^2 \log p_u(u_t)}{\partial \rho_t \partial \rho_t}\Big]\Big)^{-1/2}\nabla_t \nonumber\\ & = \frac{\sigma}{|x_{t-1}-\theta_R|}\frac{u_t}{\sigma^2}(x_{t-1}-\theta_{R}) = \sign(x_{t-1}-\theta_{R})\frac{u_t}{\sigma}.
\end{align}
\end{itemize}
The vector of parameters $\vec\lambda = [\omega, \alpha, \beta, \sigma^2]'$ is estimated with the Maximum Likelihood Estimator (MLE).

\textbf{Maximum Likelihood Estimation:} Given a set of observations $x_1,\cdots,x_T$, define $y_t = x_t-\theta_R$ for $t\in\{1,\cdots,T\}$ the likelihood for the Score-Driven model (\ref{eq21}) is
\begin{align}
    \mathcal{L}_T \vcentcolon = \mathcal{L}(\vec\lambda|y_1,\cdots,y_T) & = p(y_1)\prod_{t=2}^T p(y_t|y_{t-1};\vec\lambda).
\end{align}
The constant term $p(y_1)$ does not give an important contribution for longer and longer time series. As such, it is neglected for convenience.
Since $u_t$ is identically and independent standard normally distributed, it is
\begin{equation}
    y_t|y_{t-1}\sim\mathcal{N}(\rho_{t,d} y_{t-1},\sigma^2),
\end{equation}
hence
\begin{equation}
    p(y_t|y_{t-1}) = \frac{1}{\sqrt{2\pi\sigma^2}}\exp\Big\{-\frac{(y_t-\rho_{t,d}y_{t-1})^2}{2\sigma^2}\Big\}.
\end{equation}
The logarithm of the likelihood given the observations $y_1,\cdots,y_T$ can be stated as
\begin{align}
    \ln\mathcal{L}_T \vcentcolon = \ln\mathcal{L} (\vec\lambda|y_1,\cdots,y_T) = \sum_{t=2}^T\Big\{-\frac{1}{2}\ln(2\pi) - \frac{1}{2}\ln\sigma^2-\frac{1}{2\sigma^2}(y_t-\rho_{t,d}y_{t-1})^2\Big\}.    
%    & = \ln\prod_{t=2}^T p(y_t|y_{t-1};\vec\lambda)\nonumber\\
%    & = \sum_{t=2}^T\ln p(y_t|y_{t-1};\vec\lambda)\nonumber\\
%    & = \sum_{t=2}^T\Big\{-\frac{1}{2}\ln(2\pi) - \frac{1}{2}\ln\sigma^2-\frac{1}{2\sigma^2}(y_t-\rho_{t,d}y_{t-1})^2\Big\}.
\end{align}
The MLE is then defined as
%\begin{equation}
%    \vec\lambda_T^* = \argmax_{\vec\lambda}\ln\mathcal{L}_T
%\end{equation}
%or, equivalently, as
\begin{equation}\label{eq: MLE estimation}
    \vec\lambda_T^* = [\omega^*,\alpha^*,\beta^*,(\sigma^2)^*] =\argmin_{\vec\lambda}\Big\{-\ln\mathcal{L} _T\Big\}.
\end{equation}
We substitute the optimal value of the parameter vector to the time-varying correlation and we get 
\begin{equation}
    \rho_{t,d} = \omega^*+\alpha^* s_{t-1,d} + \beta^*\rho_{t-1,d},
\end{equation}
the UPM term results as
\begin{align}
    p(x_{t+1}|x_t^{(r_t)})
    & = \mathcal{N}\big(\mu_{r_t}+\rho_{t,d}(x_t-\mu_{r_t}),\sigma^2+\sigma_{r_t}^2\big),
\end{align}
where the posterior parameters are 
\begin{align}\label{eq: post params MBOC}
    \mu_{r_t} = \frac{b_{r_t}+\frac{\mu_0}{\sigma_0^2}}{a_{r_t}+\frac{1}{\sigma_0^2}}\quad\text{and} \quad\sigma_{r_t}^2 = \Big(a_{r_t}+\frac{1}{\sigma_0^2}\Big)^{-1}\quad\text{for}\quad r_t\in\{1,...,t\},
\end{align}
and
\begin{align}
    a_{r_t} & = \frac{1}{\sigma^2}+\frac{(r_t-1)(1-\rho_{t,d})^2}{\sigma^2(1-\rho_{t,d}^2)}\\
     b_{r_t} & = \begin{cases}\frac{x_{t}}{\sigma^2},\quad\text{for}\quad r_t=1,\\[10pt]
    \frac{x_{t-1}}{\sigma^2}+\frac{(1-\rho_{t,d})(x_t-\rho_{t,d} x_{t-1})}{\sigma^2(1-\rho_{t,d}^2)},\quad\text{for}\quad r_t=2,\\[10pt]
        \frac{x_{t+1-r_t}}{\sigma^2}+\frac{(1-\rho_{t,d})^2\sum_{i=t+2-r_t}^{t-1}x_i+(1-\rho_{t,d})(x_t-\rho_{t,d} x_{t+1-r_t})}{\sigma^2(1-\rho_{t,d}^2)},\quad\text{for}\quad r_t\in\{3,...,t\}.
    \end{cases}.
\end{align}
In algorithm \ref{alg: MBOC new}, the vector of parameters $\vec \lambda$ is estimated at each time step after we de-mean the data with the posterior mean, see Eq. (\ref{eq: post params MBOC})\footnote{In \cite{QFpaper} we used a slightly different algorithm but the results are essentially the same.}. In particular, at each time step $t>1$, we estimate the run length 
\begin{equation*}
    i = \argmax_{i\in\{1,...,t\}}p(r_t=i|x_{1:t})
\end{equation*}
and we define the {\it de-meaned data set} as $x_t^{(i)}-\mu_i = \{x_{t+1-i}-\mu_i,...,x_t-\mu_i\}$. In other words, we remove the mean associated with the regime of length $i$. For the so-obtained sub-sample of the time series data, we infer $\vec \lambda$ via MLE, and we filter $\rho_{t,d}$ using the Score-Driven model in Eq. (\ref{eq:scoredrivenrho}).

For the sake of robustness in terms of out-of-sample Mean Squared Error (MSE) (see section \ref{sec:results} for details), (i) we define a threshold value\footnote{The threshold value $\eta$ works as a hyperparameter that is tuned in a preliminary phase, see the implementation details below.} $\eta$ for a minimum regime length; then (ii) we filter the time-varying correlation; finally, (iii) we infer the variance $\sigma^2$ for each regime whenever $i>\eta$. As such, the de-meaned data set contains at least $\eta$ data points.

\subsubsection{MBOV model}\label{subsec: MBOV model}

We also extend the Gaussian auto-correlated case by promoting the variance (i.e., $\sigma^2$) to time-varying $\sigma_{G_t}^2$ within a regime by describing the evolution with a GARCH(1,1) model \cite{EngleBollerslev}, that is a special case of the Score-Driven model. We name this model as MBOV. In particular, within a regime $R$, the data-generating process is
\begin{align}\label{eq: AR-GARCH}
    x_t & = \theta_{R}(1-\rho) + \rho x_{t-1}+\sqrt{1-\rho^2}\nu_t\sigma_{G_t}\quad\text{with}\quad\sigma_{G_t}^2 = \alpha_0+\alpha_1 (1-\rho^2)(\nu_{t-1}\sigma_{G_{t-1}})^2+\beta\sigma_{G_{t-1}}^2,
\end{align}
where $\alpha_0,\alpha_1,\beta \geq 0 \quad\text{and}\quad \nu_t \sim \mathcal{N}(0,1)$. The mean $\theta_R$ and the correlation $\rho$ are assumed to be unknown across different regimes. 

At every time step, the aim is to filter the variance $\sigma_{G_t}^2$ and infer the correlation $\rho$. Eq. (\ref{eq: AR-GARCH}) can be stated as
\begin{equation}\label{eq: AR1}
    x_t-\theta_R = \rho(x_{t-1}-\theta_R)+\sqrt{1-\rho^2}\nu_t\sigma_{G_t},
\end{equation}
where we observe that the DGP is an AR(1)-GARCH(1,1) model. As such, we can infer the correlation $\rho$ in the first step by fitting an AR(1) model in the de-meaned regime. Then, we filter the variance $\sigma_{G_t}^2$ by fitting a GARCH(1,1) model for the residuals.

Given the assumption for $\rho$, Eq. (\ref{eq: AR-GARCH}) is equivalent to% to a GARCH(1,1) model with constant mean $C$:
\begin{align}
    x_t & = C+\sqrt{1-\rho^2}\nu_t\sigma_{G_t}\quad\text{with}\quad\sigma_{G_t}^2 = \alpha_0+\alpha_1 x_{t-1}^2+\beta\sigma_{G_{t-1}}^2,
\end{align}
where
\begin{equation}
    C = \theta_{R}(1-\rho) + \rho x_{t-1}.
\end{equation}
We filter the variance as in a GARCH model at every time step by considering data from the most likely regime found thus far. We set the initial condition for the GARCH(1,1) as equal to $\sigma_{G_0}^2$. 

The predictive posterior distribution is
\begin{equation}
    p(x_{t+1}|x_{t}^{(r_t)}) = \mathcal{N}(\mu_{r_t}+\rho(x_t-\mu_{r_t}),\sigma_{G_t}^2(1-\rho^2)+\sigma_{r_t}^2(1-\rho)^2),
\end{equation}
where the posterior parameters are 
\begin{align}\label{eq: post params MBOC}
    \mu_{r_t} = \frac{b_{r_t}+\frac{\mu_0}{\sigma_0^2}}{a_{r_t}+\frac{1}{\sigma_0^2}}\quad\text{and} \quad\sigma_{r_t}^2 = \Big(a_{r_t}+\frac{1}{\sigma_0^2}\Big)^{-1}\quad\text{for}\quad r_t\in\{1,...,t\},
\end{align}
and
\begin{align}
    a_{r_t} & = \frac{1}{\sigma_{G_0}^2}+\frac{(r_t-1)(1-\rho)^2}{\sigma_{G_t}^2(1-\rho^2)},\\
    b_{r_t} & = \begin{cases}\frac{x_{t}}{\sigma_{G_0}^2},\quad\text{for}\quad r_t=1,\\[10pt] 
    \frac{x_{t-1}}{\sigma_{G_0}^2}+\frac{(1-\rho)(x_t-\rho x_{t-1})}{\sigma_{G_t}^2(1-\rho^2)},\quad\text{for}\quad r_t=2,\\[10pt]
        \frac{x_{t+1-r_t}}{\sigma_{G_0}^2}+\frac{(1-\rho)^2\sum_{i=t+2-r_t}^{t-1}x_i+(1-\rho)(x_t-\rho x_{t+1-r_t})}{\sigma_{G_t}^2(1-\rho^2)},\quad\text{for}\quad r_t\in\{3,...,t\}.
    \end{cases}.
\end{align}

\begin{algorithm}[tb]
\caption{MBOV}
\label{alg: MBOV}
\textbf{Input}: $\mu_0,\sigma_0^2, \sigma_{G_0}^2, \sigma^2, \rho, p(r_0=0) = 1$\\
\textbf{Output}: $p(r_t|x_{1:t}), \hat{\mu}_t$
\begin{algorithmic}[1] 
\For{$t = 1,...$} 
\State Observe $x_t\sim\mathcal{N}(\theta_R,\sigma^2)$
\State Compute $\hat{\mu}_t$
\State Find $\text{argmax}_{i\in\{0,1,...,t\}}p(r_t=i|x_{1:t})$
\State Remove the mean from $x_{t}^{(i)}$ and get the data $y_t^{(i)}$
\Comment{$y_t^{(i)}$ is the de-meaned data of $x_t^{(i)}$}
\If{$t>\eta$}
\State Infer $\rho$ with AR(1) (Eq. (\ref{eq: AR1})) using the data $y_t^{(t)}$
\State Filter $\sigma_{G_t}^2$ with GARCH(1,1) on the residuals of Eq. 
 (\ref{eq: AR1})
\EndIf
\State Compute $p(r_t|x_{1:t})$
\State Update $\mu_t,\sigma_t^2$
\EndFor
\end{algorithmic}
\end{algorithm}
See Algorithm \ref{alg: MBOV} for the details on the inference procedure.
\clearpage
%====================================
%============ Results ===============
\section{Simulations}
\label{sec:results}

Before applying the proposed models MBO($q$) and Score-Driven MBO(1) to empirical data, we test their performance on simulated data by comparing them with the baseline method in terms of CPD and forecasting accuracy. The analysis points out that utilizing MBO(q) models to capture autocorrelation in time series is crucial for both forecasting accuracy and CPD, even when the parameters are misspecified (section \ref{subsec:BOCPDvsMBO(1)}). We show that the Score-Driven model captures the true dynamics of the correlation. In the case of regimes for the mean, the Score-Driven AR(1) model (without regimes) overestimates the autocorrelation, while MBOC better captures the underlying dynamics (section \ref{subsec:Filtering autocorrelation with MBOC}). We finally corroborate the maximum likelihood estimation of the MBOC model with simulations by comparing MBOC estimates with the ones based on the Score-Driven AR(1) model (section \ref{sec: Perf of MLE}).\footnote{Similar results hold for the MBOV model and are not shown here for reasons of space. However, they are available upon request.}

%=================================
\subsection{Comparison between BOCPD and MBO(1)}\label{subsec:BOCPDvsMBO(1)}
In this section, we simulate dependent data $x_{1:T}$ with $T=200$ of order $q=1$ with regimes in the mean, we will examine the performance of the BOCPD versus the MBO(1) model in terms of their forecasting accuracy and it terms of change-point detection. We simulate the arrival of change-points as a Bernoulli process with probability of success equal to $1/70$. Each time a new change-point arrives we sample the mean of the new regime as:  $\theta_R\sim\mathcal{N}(0,5)$ and we assume that the DGP within each regime to be:
\begin{align}\label{eq: DGP for MBOC}
    x_t & \sim \mathcal{N}(\theta_R,2)\\
    x_t|x_{t-1} & \sim\mathcal{N}(\rho(x_{t-1}-\theta_R)+\theta_R,2(1-\rho^2)).
\end{align} 
Three different simulating scenarios are created according to the value of $\rho\in\{0.1,0.4,0.7\}$. We perform 100 simulations for each scenario. We compare the two models by setting $\mu_0=0, \sigma^2_0 = 2, \sigma^2 = 2$ and $1/h = 1/70$ (the hazard rate) for both models and $\rho = 0.4$ for the MBO(1) model. 

The forecasting performances are compared by computing the MSE for the predictive mean of each model. The predictive mean $\hat{\mu}_t$ at time $t$ is the one-step-ahead forecast given by
\begin{equation}\label{eq: pred mean}
    \hat{\mu}_{t} = \sum_{r_t} p(x_{t+1}|x_{1:t},r_t,)p(r_t|x_{1:t})=\sum_{r_t} \mu_{r_t} p(r_t|x_{1:t})
\end{equation}
where $\mu_{r_t}$ is defined in Eq.(\ref{eq: post params BOCPD}) for the BOCPD and in Eq.(\ref{eq: post params MBO1}) for the MBO(1). The MSE is:
\begin{equation}\label{eq: MSE}
    \text{MSE} = \frac{1}{T}\sum_{t=1}^T(\hat{\mu}_{t-1}-x_t)^2.
\end{equation}

As a second step, we compare the two models regarding change-point detection by using the covering metric as defined in \cite{covering_metric}.
\begin{definition}
    Let $\mathcal{G}$ be the ground truth partition of $[1,T]$ and $\mathcal{G}_\text{a}$ be the partition provided by the model. Then, the covering metric is defined as
    \begin{equation}
        C(\mathcal{G},\mathcal{G}_\text{a}) = \frac{1}{T}\sum_{\mathcal{A}\in\mathcal{G}}|\mathcal{A}|\max_{A^\prime\in\mathcal{G}_\text{a}} J(\mathcal{A},\mathcal{A^\prime})
    \end{equation}
    where $J(\mathcal{A},\mathcal{A^\prime}) = \frac{|\mathcal{A}\cap\mathcal{A^\prime}|}{|\mathcal{A}\cup\mathcal{A^\prime}|}$ is the Jaccard index and $|\mathcal{A}|$ is the cardinality of set $\mathcal{A}$\footnote{Notice that $\mathcal{A}$ is a subset of $[1,T]\cap\N$.}. 
\end{definition}
The covering metric is a weighted average of the Jaccard index between ground truth intervals of regimes and intervals identified by the detection model within $[1,T]$, with weights $|\mathcal{A}|$ for $\mathcal{A}\in\mathcal{G}$ such that $\sum_{\mathcal{A}\in\mathcal{G}}|\mathcal{A}| = T$. Since $J(\mathcal{A},\mathcal{A^\prime}) \in [0,1]$, it holds that $C(\mathcal{G},\mathcal{G}_\text{a}) \in [0,1]$, with values closer to one signaling that the detection model finds regimes similar to the ground truth. The covering metric can also be used for the analysis of false positives. Consider, for example, a stationary time series $x_{1:T}$ (with no CPs) such that $\mathcal{G} = \{[1,T]\}$. Any detected CP is then a false positive, for example, $s\in[1,T/4]$ such that $\mathcal{G}_a = \{[1,s-1],[s,T]\}$. In this example, the covering metric is
\begin{equation}
    C(\mathcal{G},\mathcal{G}_a) = \frac{1}{T}T\frac{|[s,T]|}{T} = \frac{T-s+1}{T},
\end{equation}
smaller than the maximum allowed value. However, the closer $s$ is to 1, the closer the covering metric is to 1. As such, the larger is the overlap between detected regimes and ground truth, the closer is the covering metric to the maximum.

Table \ref{tab: Sim data MSE and CM} contains the results of the comparison between the two models in terms of MSE and the covering metric. Notice that, since our method is online, it is also {\it de facto} out of sample. We report the mean and the standard error for both the MSE and the covering metric over 100 simulations in the three scenarios for autocorrelation. In the first case of autocorrelated time series with $\rho=0.1$, both models are misspecified (BOCPD assuming i.i.d. data while MBO(1) assuming a Markovian autoregressive structure with autocorrelation equal to $0.4$), BOCPD outperforms MBO(1) since the effect of autocorrelation appears as negligible in terms of MSE and covering. However, when autocorrelation of data increases to $\rho=0.4$ or $\rho=0.7$, we observe a switch with MBO(1) outperforming the benchmark thanks to the better description of temporal dependencies. 
%the BOCPD has the best performance in both metrics since the autocorrelation in that model is assumed to be zero while in the MBO(1) to be 0.4. On the other hand when the dataset is more autocorrelated, with $\rho=0.4$ and $0.7$ the MBO(1) model with autocorrelation $0.4$ outperforms the baseline model. 
This result signals that capturing autocorrelation in time series using MBO($q$) models is key both in terms of forecasting and CPD, even in the case of misspecified parameters. Notice that the differences reported in Table \ref{tab: Sim data MSE and CM} are statistically significant at a $1\%$ level after performing a paired t-test.
\begin{center}
\begin{tabular}{c c||c c c}
\cline{1-5}
Comparison &$\rho$ & $0.1$ & $0.4$ & $0.7$ \\
\hline\hline
\multirow{2}{2em}{MSE}& BOCPD & \textbf{5.32 (1.60)} & 4.21 (1.5) & 2.6 (1.23)\\

 & MBO(1) & 5.62 (0.14) & \textbf{3.71 (0.12)} & \textbf{1.95 (0.09)} \\
  
  % & \textbf{MBOC} & \textbf{0.890} & \textbf{0.890} & \textbf{0.890}\\  
  \hline
\multirow{2}{1.5em}{CM} & BOCPD & \textbf{0.63 (0.17)} & 0.65 (0.2) & 0.74 (0.2)\\

 & MBO(1) & 0.58 (0.01) & \textbf{0.69 (0.19)} & \textbf{0.78 (0.01)} \\
  
 % &  \textbf{MBOC} & \textbf{0.890} & \textbf{0.890} & \textbf{0.890}\\  
  \hline
\end{tabular}
\captionof{table}{\textbf{Top}: Comparison of one-step-ahead MSE between BOCPD and MBO(1). \textbf{Bottom:} Comparison of the covering metric (CM) between BOCPD and MBO(1). \textbf{Note:} The reported MSE and CM is the mean (standard error) of the MSEs ans CMs respectively after 100 simulations. The correlation $\rho$ is the one used to simulate the data. The MBO(1) in all three cases has $0.4$ correlation.}
\label{tab: Sim data MSE and CM}
\end{center}

\subsection{Filtering autocorrelation dynamics with MBOC}\label{subsec:Filtering autocorrelation with MBOC}
In this section, we analyze the filtering properties of the Score-Driven model in a simulation setting with and without CPs in the mean by examining the MBOC performance in recovering the true dynamics of the autocorrelation\footnote{Notice that the case of time-varying variance analyzed with MBOV is similar, and it is not reported here for the sake of simplicity.} also when the actual dynamics is not the one of the model. It is well known that Score-Driven models works nicely as filters of a misspecicied dynamics. This means that they are able to estimate an unobservable time-varying parameter also when it is not following a Score-Driven process. To this end we consider a time series of length $T=1000$ generated as
\begin{align}\label{eq:simulation1}
    x_t = \rho_t(x_{t-1}-\mu) +\mu + u_t,\quad u_t\sim\mathcal{N}(0,\sigma^2)
\end{align}
with $\sigma^2=1$ and time-varying $\rho_t$. We analyze two scenarios: (i) stationary time series with no regimes and $\mu=0$ at all times, and (ii) the nonstationary case with regimes modeled by CPs as an i.i.d. Bernoulli process with probability $1/70$ and $\mu \sim \mathcal{N}(3,2)$ in each regime. For nonstationary time series, any time a CP occurs defining a new regime $R_i$, then the first realization of $R_i$ is assumed to be an independent random variable with unconditional probability distribution $\mathcal{N}(\mu_{R_i},\sigma^2)$.
%In the case when we are interested in the existence of regimes in the mean we set $\sigma^2 = 1$ and $\mu\sim\mathcal{N}(3,2)$ every time a new change-point arrives. The arrival of change-points is modelled as a Bernoulli process with probability of arrival $1/70$. On the contrary, when there are no regimes we set $\mu=0$.

\begin{figure}[!h]
    \centering
    \subfigure[]{\includegraphics[width=0.45\textwidth]{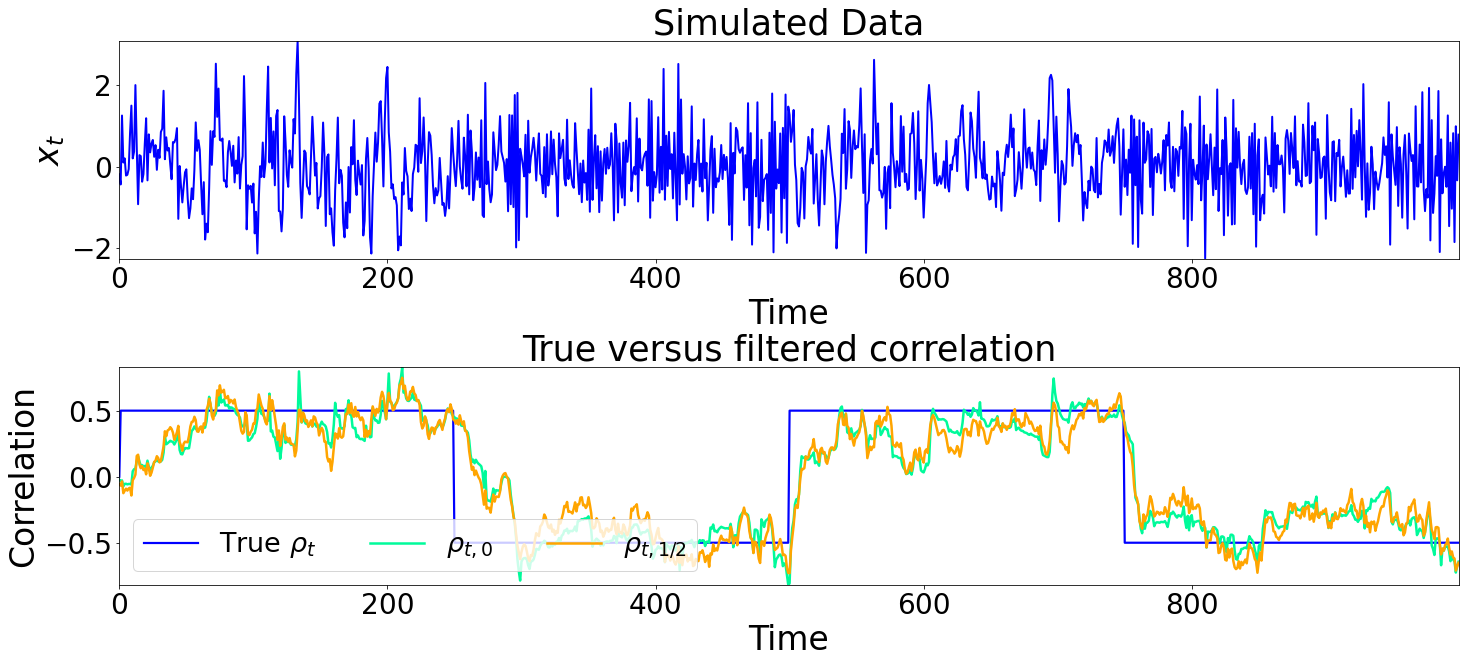}}  
    \subfigure[]{\includegraphics[width=0.45\textwidth]{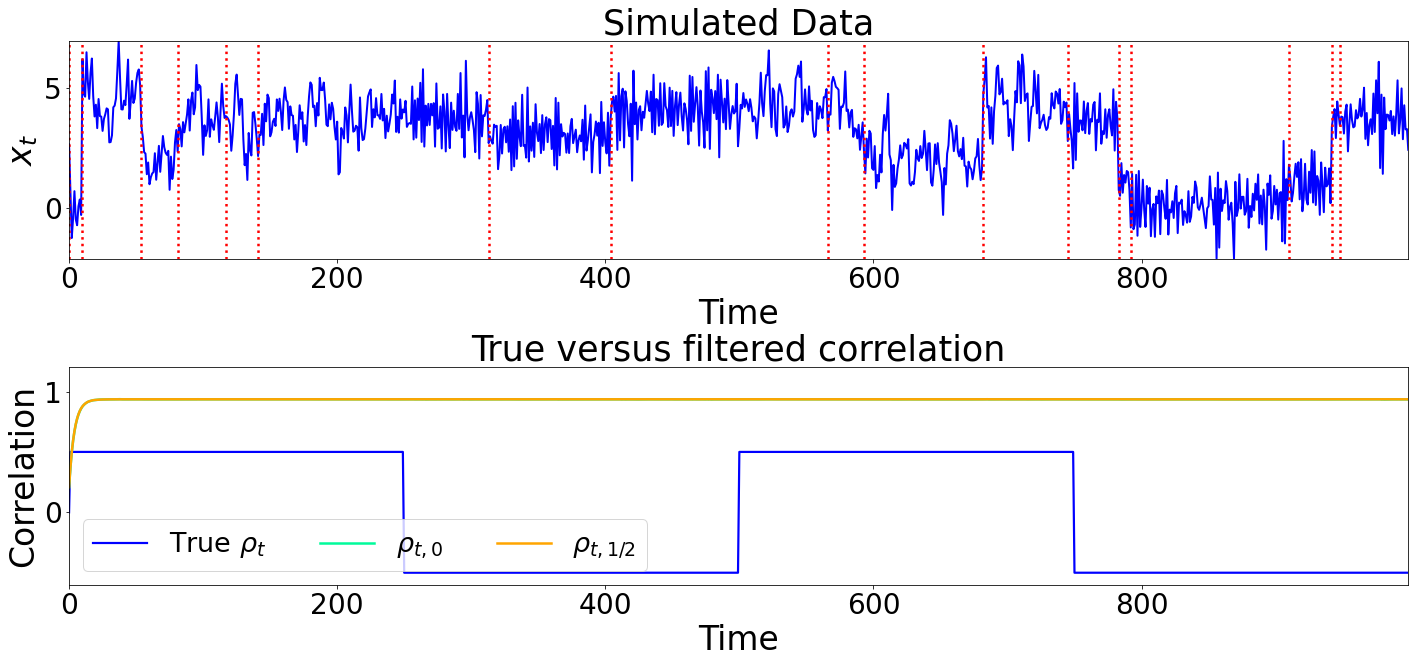}}
    \subfigure[]{\includegraphics[width=0.45\textwidth]{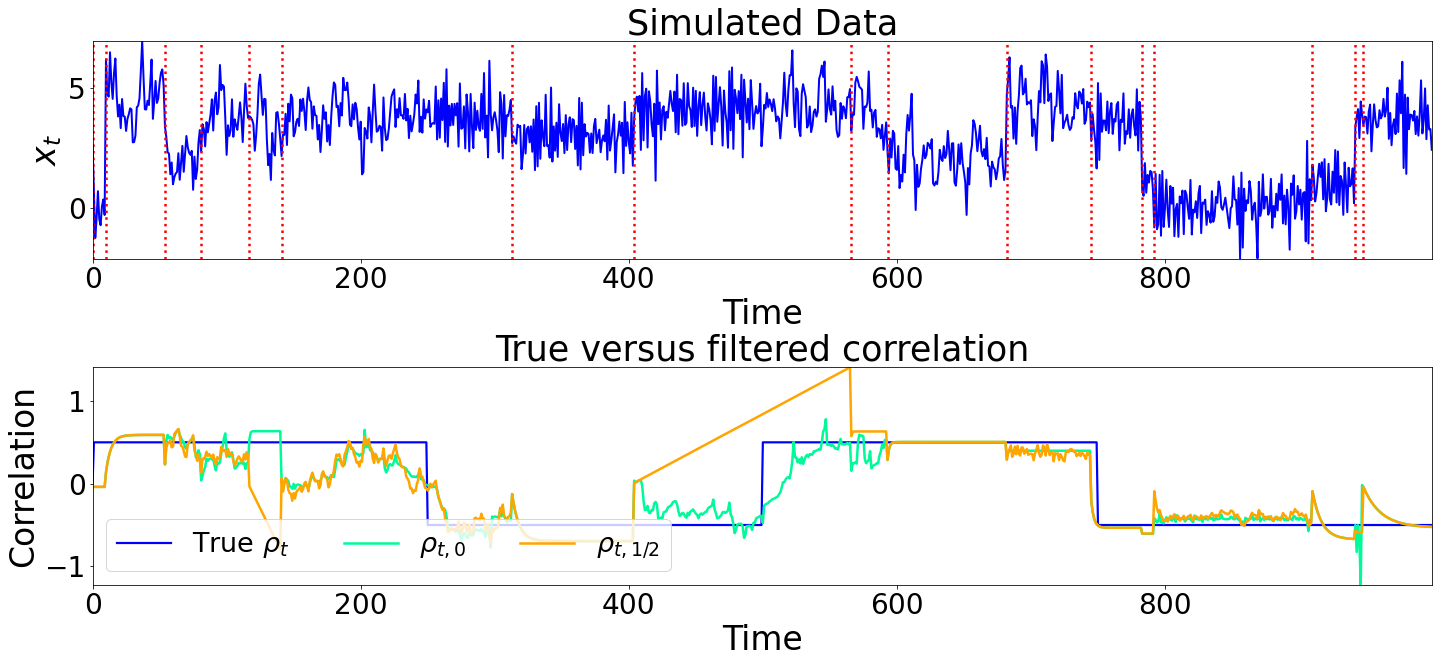}}
    \subfigure[]{\includegraphics[width=0.45\textwidth]{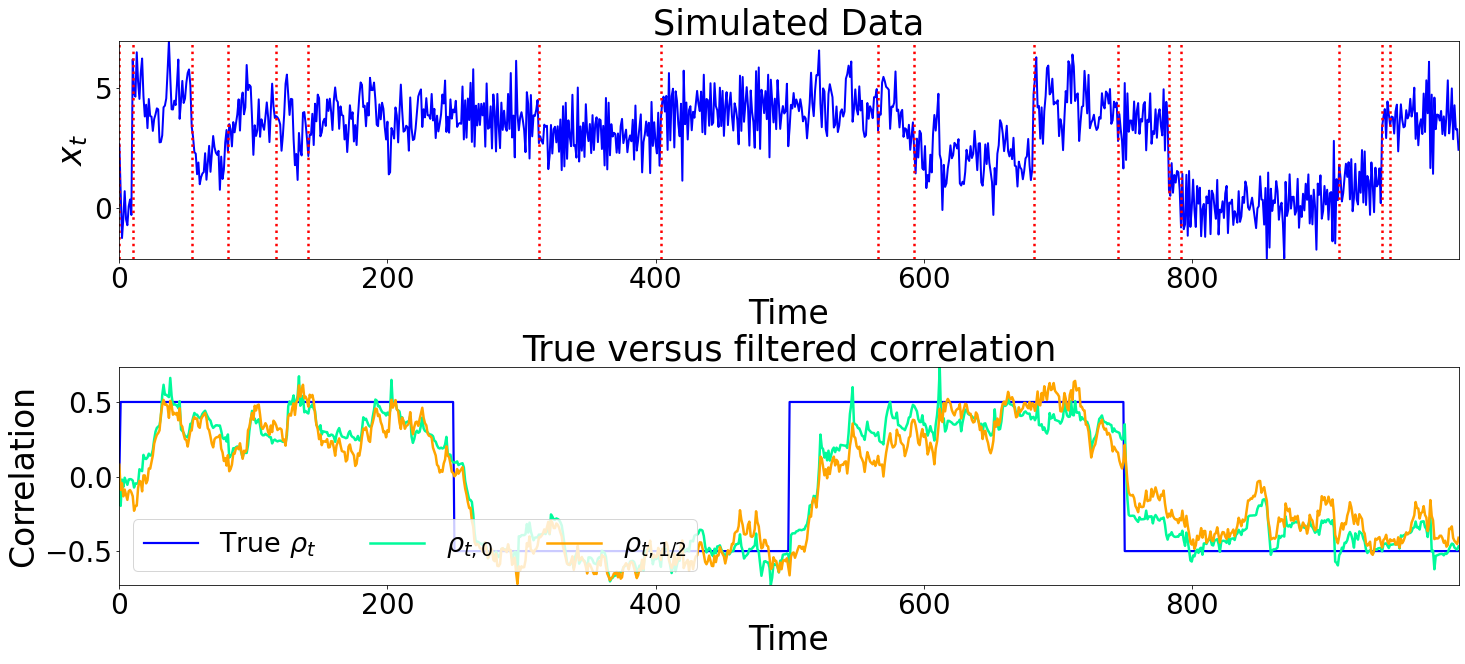}}
    \caption{\textbf{Top:} Simulation of data according to Eq.(\ref{eq:simulation1}) when the autocorrelation follows a a piecewise constant function (a) without CPs in the mean (b)-(d) with CPs in the mean. \textbf{Bottom:} The true autocorrelation as in Eq.(\ref{eq: step function}) along with the filtered correlation $\rho_{t,d}$ when the optimization is performed (a)-(b) using the original Score-Driven AR(1) model (c) using sub-samples of data from the last CP (d) when the mean is removed consistently with the detected regimes.}
    \label{fig: Filt vs True Step}
\end{figure}

\begin{figure}[!h]
    \centering
    \subfigure[]{\includegraphics[width=0.45\textwidth]{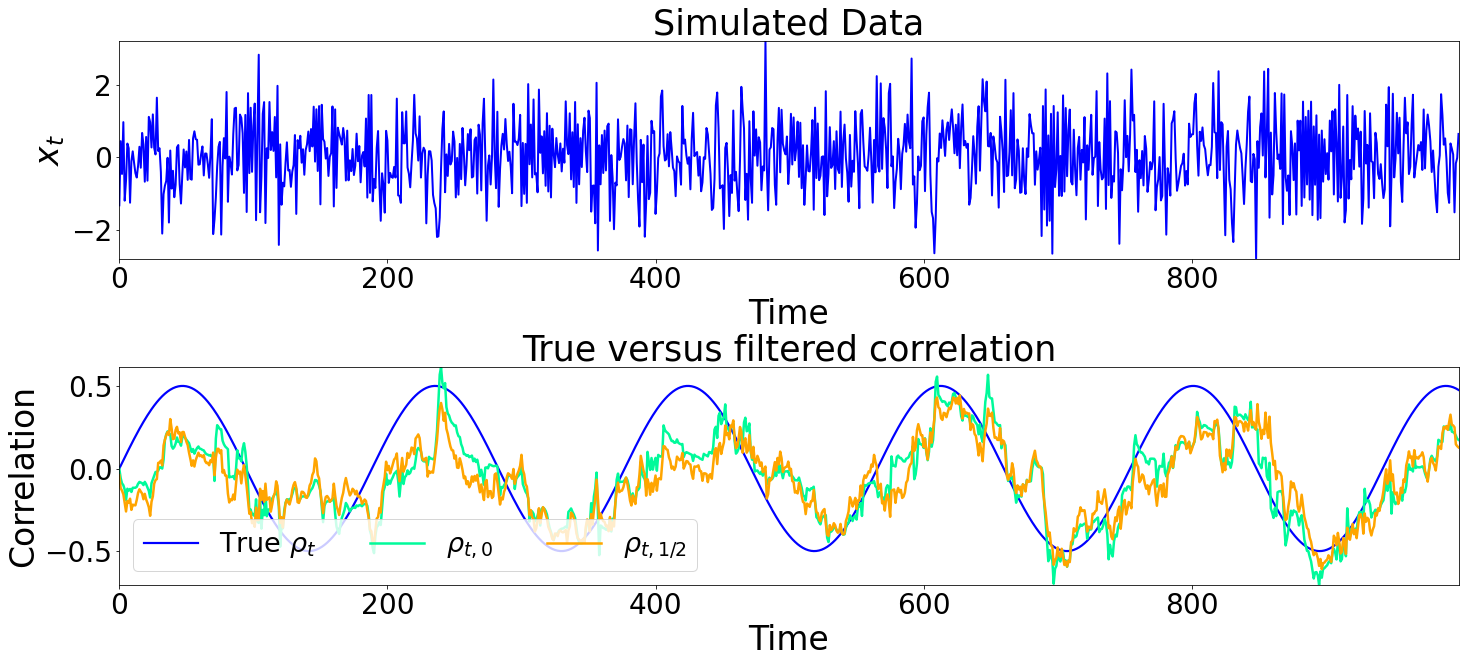}}  
    \subfigure[]{\includegraphics[width=0.45\textwidth]{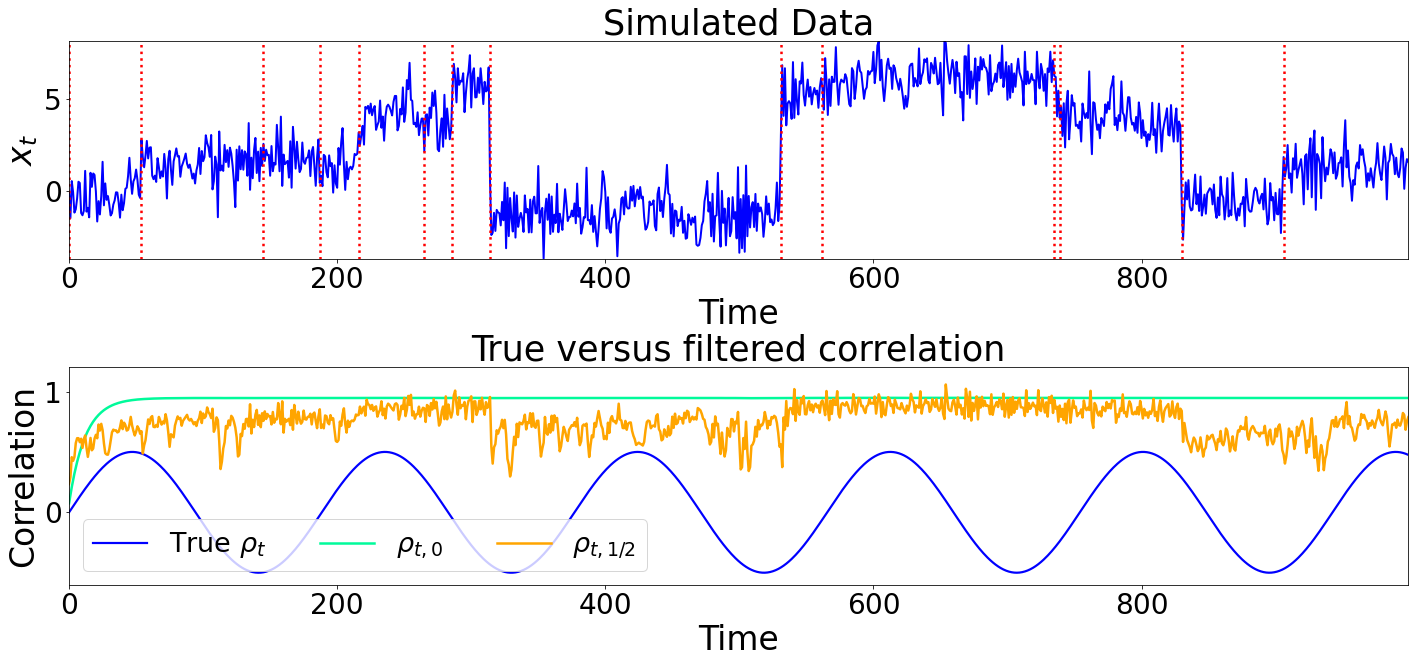}}
    \subfigure[]{\includegraphics[width=0.45\textwidth]{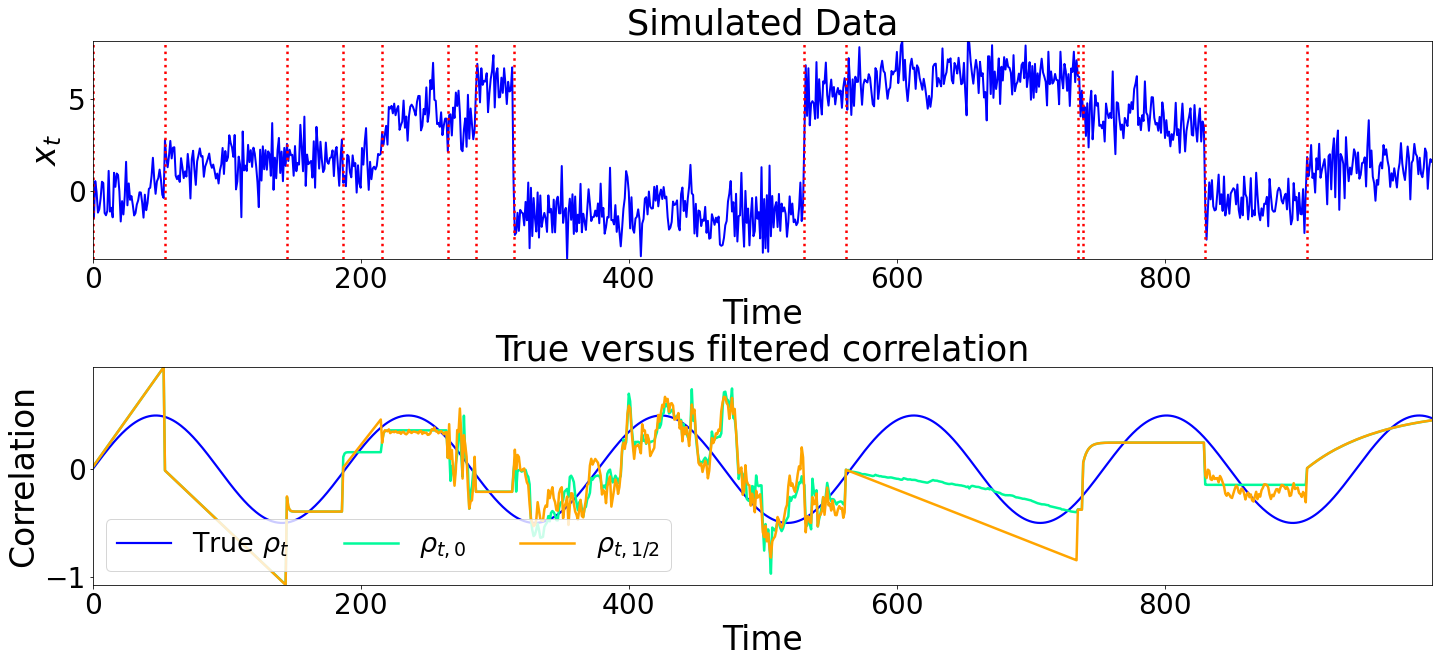}}
    \subfigure[]{\includegraphics[width=0.45\textwidth]{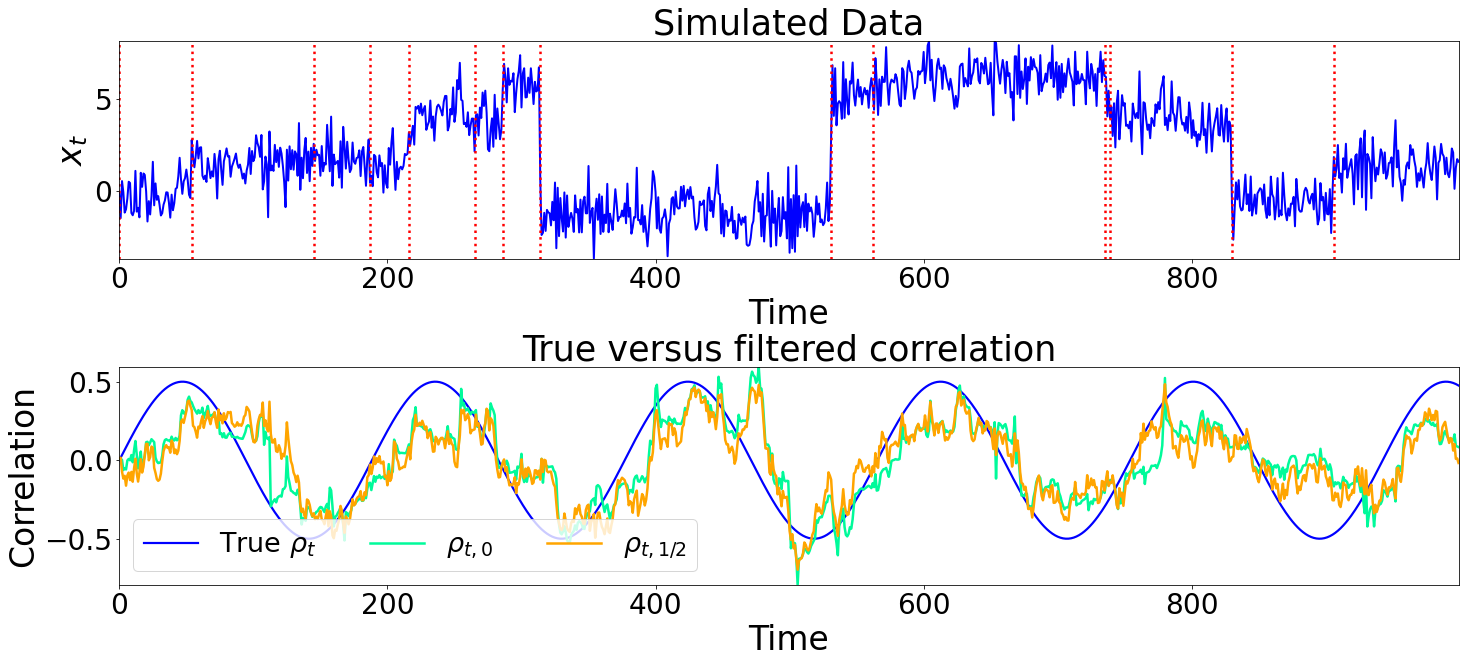}}
    \caption{\textbf{Top:} Simulation of data according to Eq.(\ref{eq:simulation1}) when the autocorrelation is a sinusoidal function (a) without CPs in the mean (b)-(d) with CPs in the mean. \textbf{Bottom:} The true autocorrelation as in Eq.(\ref{eq: sinusoid}) along with the filtered correlation $\rho_{t,d}$ when the optimization is performed (a)-(b) using the original Score-Driven AR(1) model (c) using sub-samples of data from the last CP (d) when the mean is removed consistently with the detected regimes.}
    \label{fig: Filt vs True Sin}
\end{figure}

We consider two cases for the time-varying pattern followed by $\rho_t$, namely a piecewise constant evolution,
    \begin{equation}\label{eq: step function}
        \rho_t = \begin{cases}
        0.5,\quad t<250\\
        -0.5,\quad 250\leq t<500\\
        0.5,\quad  500\leq t< 750\\
        -0.5,\quad t\geq 750
        \end{cases}
    \end{equation}
and a sinusoidal signal 
    \begin{equation}\label{eq: sinusoid}
        \rho_t = 
            \frac{1}{2}\sin{\Big(\frac{t}{30}\Big)}.
    \end{equation}
The results for the two cases are summarized in Figure \ref{fig: Filt vs True Step} and Figure \ref{fig: Filt vs True Sin}, respectively. Each figure reports four panels one for each of the inferential processes we describe below. Then, each panel shows two subpanels, namely a sample realization of the time series process (top) and the corresponding pattern of autocorrelation (bottom).

%The experiments related to simulated data when the true correlation is the step function are gathered in figure \ref{fig: Filt vs True Step} while for the sinusoid in figure \ref{fig: Filt vs True Sin}. 

First, we consider stationary time series with no regimes for the mean for both piecewise and sinusoidal time-varying autocorrelation, see panel (a) of Figures \ref{fig: Filt vs True Step} and \ref{fig: Filt vs True Sin}, respectively. The autocorrelation pattern (blue) can be consistently estimated via the Score-Driven version of the AR(1) model introduced by \cite{Score-Driven_AR} (with no regimes), and the results are in line with the one obtained by the authors. Throughout all the cases displayed in the figures, we show the results for the filtering process of the autocorrelation for both $d=0$ (green) and $d=1/2$ (orange) in Eq. (\ref{eq: I}). However, the dynamics of the autocorrelation is {\it not} captured by the original Score-Driven AR(1) model when data are generated with regimes, considering both piecewise and sinusoidal patterns, see panels (b) in the figures. In particular, autocorrelation is systematically overestimated by the model of \cite{Score-Driven_AR}. Such a result is in line with \cite{r:01,MikoschStarica2004}, where the authors show that long-range dependencies can arise because of non-stationarity. In other words, regimes for the mean of the time series result in an overestimation of the autocorrelation when the model does {\it not} account for CPs.

The proposed MBOC approach can be used to overcome this issue. In the first instance, the model in Eq. (\ref{eq:simulation1}) is estimated via MLE considering the sub-sample of data from the last CP. The results are shown in panels (c) of the figures. The autocorrelation is not overestimated anymore; nevertheless, the inference is noisy because of a lack of data, especially for short regimes. Under the assumption of stationarity of the process, once the mean is removed consistently with the detected regimes, the autocorrelation pattern can be estimated as explained above, see Algorithm \ref{alg: MBOC new}. For the latter case, the results are shown in panels (d) of the figures, displaying a superior performance with respect to the previous cases.
\clearpage
%================================
\subsection{Maximum likelihood estimation of the MBOC model}\label{sec: Perf of MLE}
In a manner similar to the previous section, we analyze the performance of the MLE $\vec\lambda_T^*$ (see Eq.(\ref{eq: MLE estimation})) under the simulation setting with CPs in the mean. This time, however, we assume that the unobservable time-varying parameter follows a Score-Driven process. The time series is of length $T = 500$, generated according to Eq.(\ref{eq21})-(\ref{eq:scoredrivenrho}) and for the time-varying correlation we set $\omega=0.001,\alpha=0.1,\beta=0.9,\sigma^2=1$. For each simulation we estimate the parameter vector $\vec\lambda^*=[\omega^*,\alpha^*,\beta^*,(\sigma^2)^*]$ and we compute the relative error for each hyperparameter, namely 
\begin{equation}
    \frac{\omega^*-\omega}{\omega},\quad\frac{\alpha^*-\alpha}{\alpha},\quad\frac{\beta^*-\beta}{\beta},\quad\frac{(\sigma^2)^*-\sigma^2}{\sigma^2}.
\end{equation}
%and we estimate the corresponding density after $N=100$ simulations.
\begin{figure}[!h]
    \centering
    \subfigure[]{\includegraphics[width=0.48\textwidth]{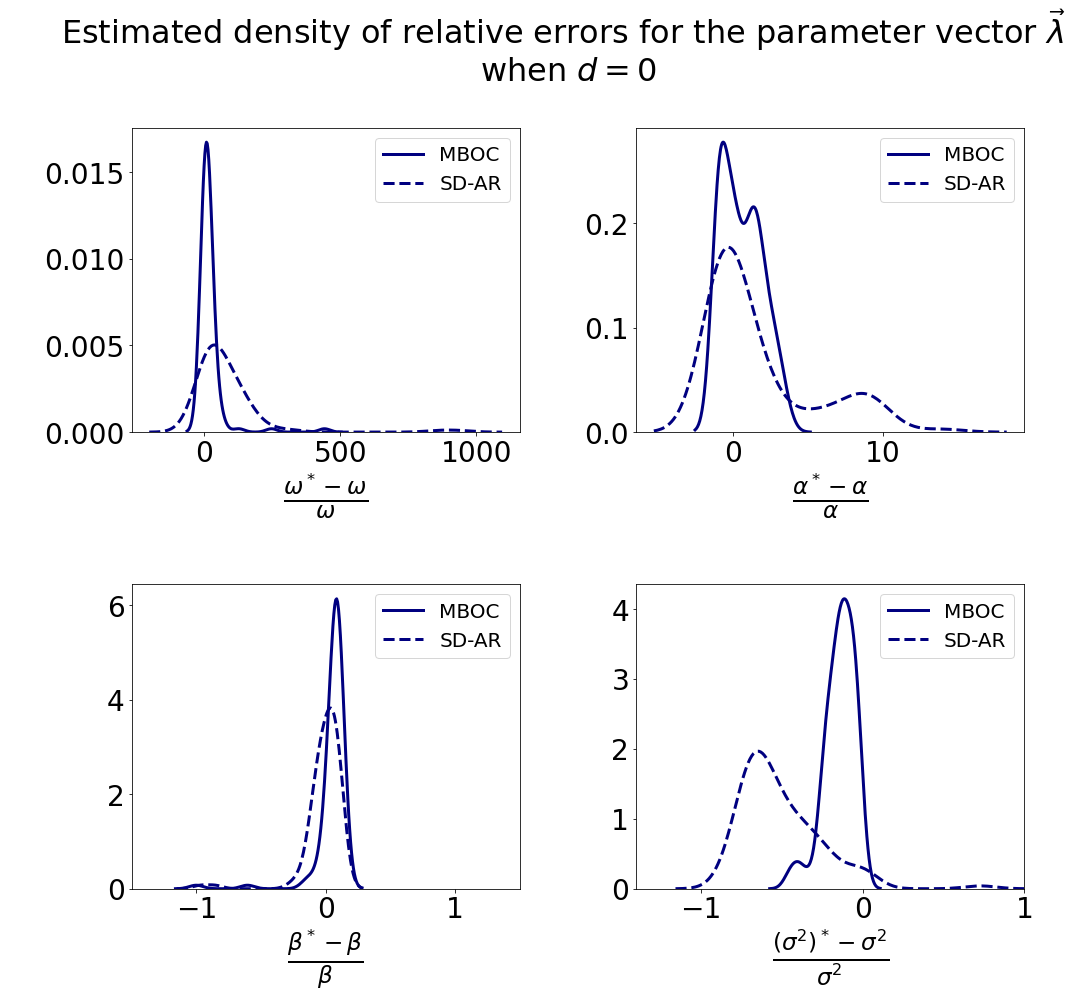}}  
    \subfigure[]{\includegraphics[width=0.48\textwidth]{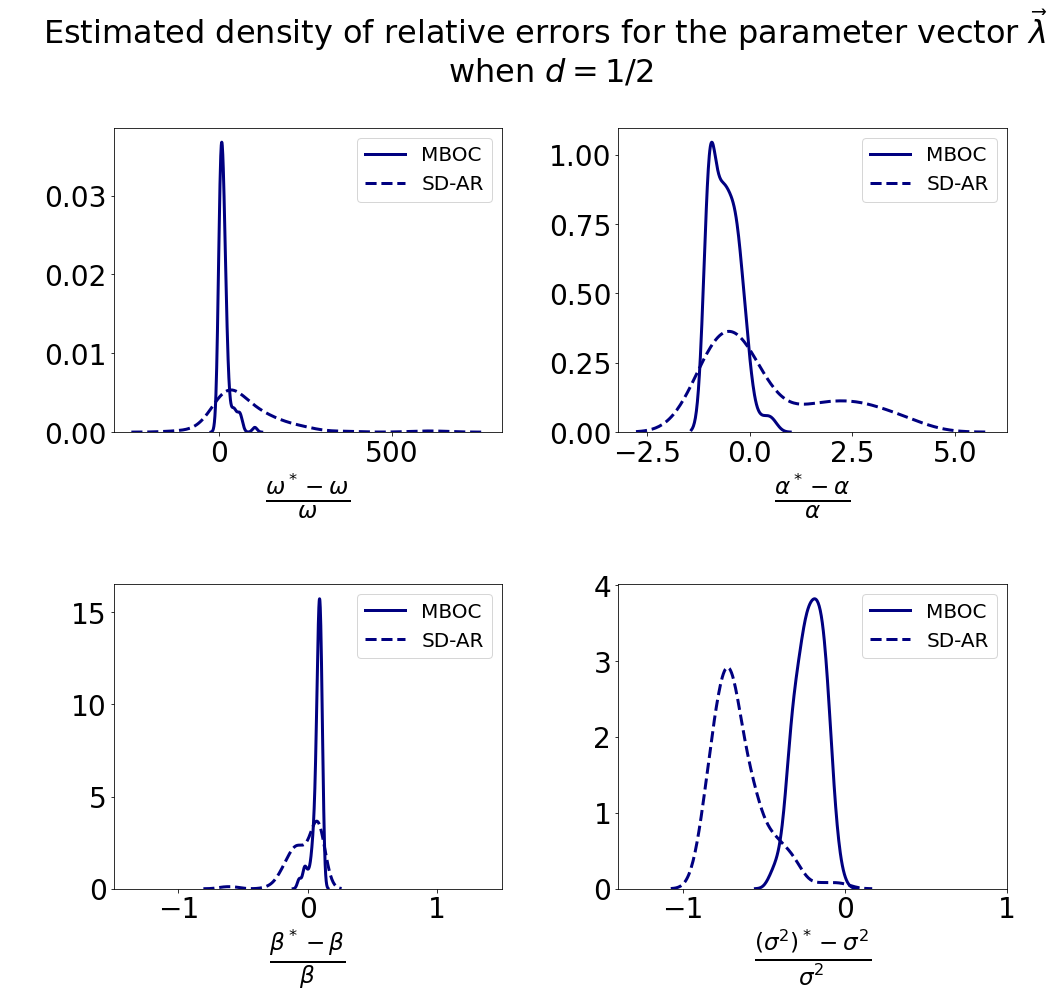}}
    \caption{Density estimation of the relative errors of the static parameters in Eq.(\ref{eq:scoredrivenrho}) (a) when $d=0$ and (b) when $d=1/2$.}
    \label{fig: density estimation}
\end{figure}
Figure \ref{fig: density estimation} shows the estimated density of the relative errors of the static parameters in Eq.(\ref{eq:scoredrivenrho}) based on $N=100$ simulations, when $d=0$ (a), and $d=1/2$ (b). The blue solid line is the estimated density for the MBOC, while the dashed blue line is the density for the original Score-Driven AR(1) model (SD-AR). We observe that in all cases, the estimated distributions for the MBOC are more concentrated around zero than those from the SD-AR. As such, the results demonstrate that the proposed methodology captures the true dynamics of the unobserved variable better than the standard Score-Driven model when there are CPs in the mean.

%==================================
\section{Empirical Applications}\label{sec:empirical}

In this Section, we apply the methods presented in Section \ref{sec:methods} to real data and we compare them in terms of their forecasting accuracy.
To compare the statistical significance of the difference in forecasting accuracy between models, we use the Diebold-Mariano (DM) test \cite{DM_test}. The DM test determines whether two forecasts are significantly different by comparing their residuals. Assuming that the actual time series is $x_{1:t}$ and the two forecasts are $\hat{x}_{1:t}$ and $\Tilde{x}_{1:t}$ the residuals for each forecast are defined as
\begin{equation}
    \hat{e}_i = \hat{x}_i-x_i\quad\text{and}\quad \Tilde{e}_i = \Tilde{x}_i-x_i\quad\text{for}\quad i=1,\cdots,t
\end{equation}
respectively. By choosing a loss function to compare the residuals, the so-called time series of loss-differential it is then defined. Frequently used loss functions are:
\begin{itemize}
\item The Mean Absolute Difference (MAD)
\begin{equation}\label{eq: mad}
    d_i = |\hat{e}_i|-|\Tilde{e}_i|\quad\text{for}\quad i=1,\cdots,t.
\end{equation}
    \item The polynomial \begin{equation}\label{eq: poly}
    d_i = \hat{e}_i^p-\Tilde{e}_i^p\quad\text{for}\quad i=1,\cdots,t
\end{equation}
when $p=2$ is the usual MSE.
\end{itemize}
The DM statistic is then defined as
\begin{equation}
    \text{DM} = \frac{\Bar{d}}{\sqrt{[\gamma_0+2\sum_{k=1}^{h-1}\gamma_k]/T}}
\end{equation}
where
\begin{equation}
    \Bar{d} = \frac{1}{T}\sum_{i=1}^T d_i\quad\text{and}\quad \gamma_k = \frac{1}{T}\sum_{i=k+1}^{T}(d_i-\Bar{d})(d_{i-k}-\Bar{d}).
\end{equation}
$h$ is the variable that quantifies the number of steps ahead the prediction is performed. In the rest of the applications we use $h=1$. 

Under the null hypothesis that the two models are equivalent in forecasting, (i.e. $\E[d_i] = 0$) the DM statistic follows a standard normal distribution $\text{DM}\sim\mathcal{N}(0,1)$.

%================================

\subsection{Bee Waggle Dance}
\begin{figure}[t]
\centering
\includegraphics[width=1\columnwidth]{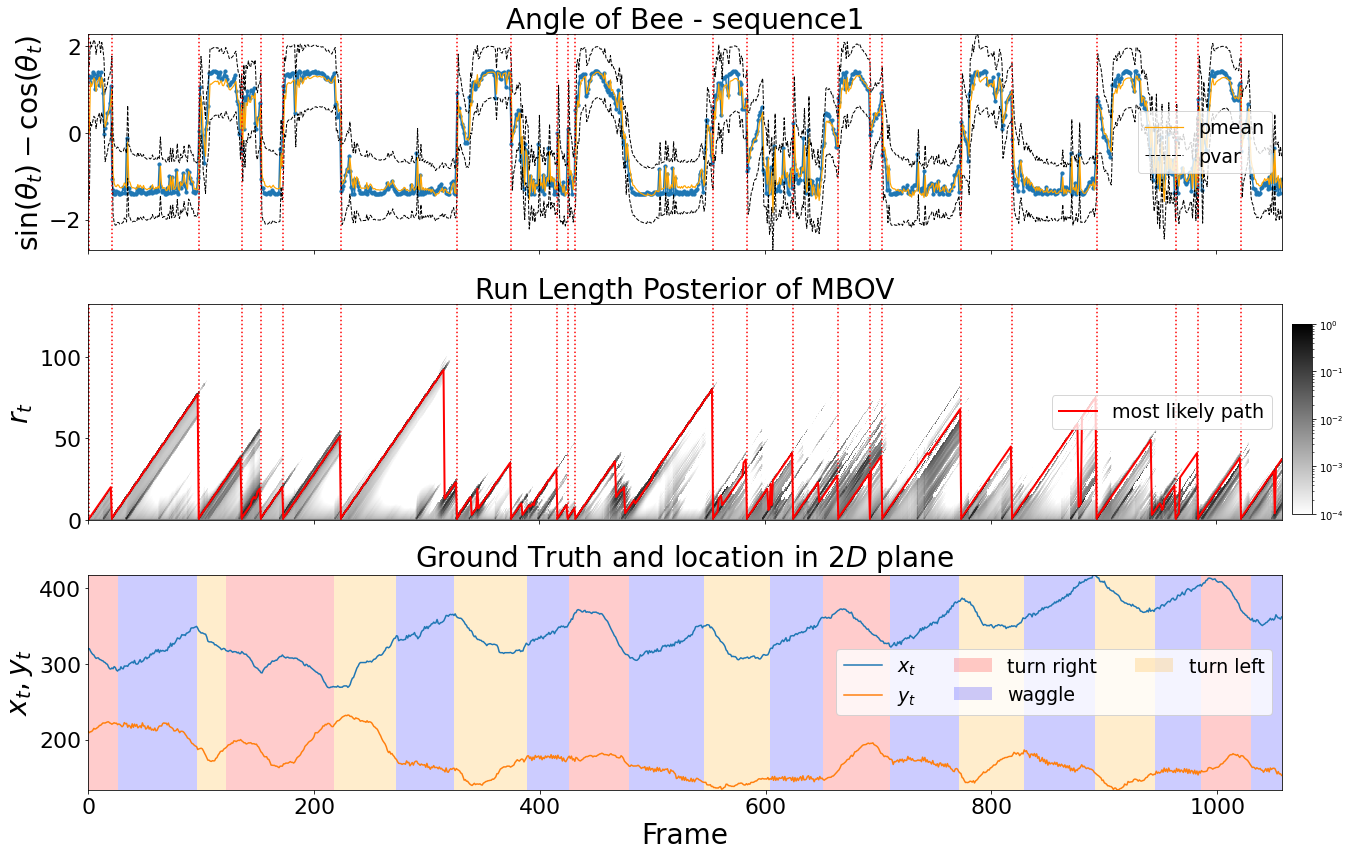}
\caption{\textbf{Top:} The difference of the sine and cosine function of the angle $\theta$ of the bee from sequence 1 as a function of video frame. In yellow the predictive mean and the dashed line is the predictive variance. \textbf{Middle:} The run length posterior of MBOV. The red line the most likely path i.e. value $r_t$ with the largest run length posterior $p(r_t|x_{1:t})$ for each $t$. \textbf{Bottom:} The ground truth labeling of the bee dance. In blue the trajectory of the $x$-axis and in orange the trajectory of the $y$-axis.}
\label{fig: MBOC_seq1}
\end{figure}

% \begin{figure}[t]
% \centering
% \includegraphics[width=0.5\columnwidth]{ACF_Hist_Bees_seq1.png}
% \caption{\textbf{Left:} The Autocorrelation plot for the angle of the bee from sequence 1. \textbf{Right:} The Histogram plot for the angle difference of the bee.}
% \label{fig: ACF_Hist_Bees_seq1}
% \end{figure}

% \begin{figure}[t]
% \centering
% \includegraphics[width=0.5\columnwidth]{acf_bees_angle_diff.png}
% \caption{\textbf{Left:} The Autocorrelation plot for the angle difference of the bee from sequence 1. \textbf{Right:} The Histogram plot for the angle of the bee.}
% \label{fig: acf_Hist_Bees_seq1}
% \end{figure}

\begin{figure}[!h]
    \centering
    \subfigure[]{\includegraphics[width=0.22\textwidth]{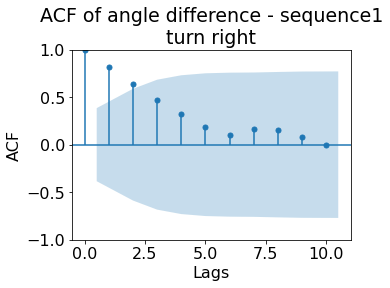}}  
    \subfigure[]{\includegraphics[width=0.22\textwidth]{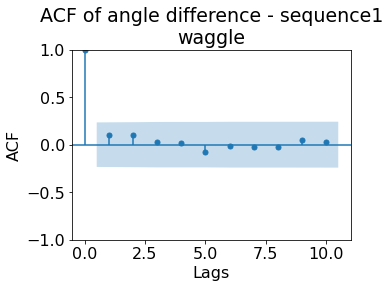}}
    \subfigure[]{\includegraphics[width=0.22\textwidth]{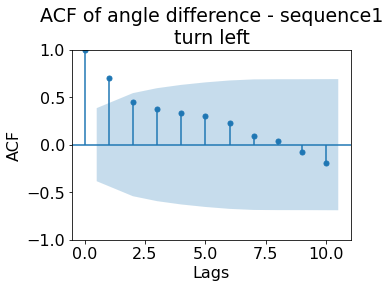}}  
    \subfigure[]{\includegraphics[width=0.22\textwidth]{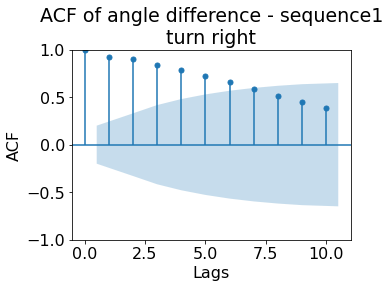}}
    \subfigure[]{\includegraphics[width=0.22\textwidth]{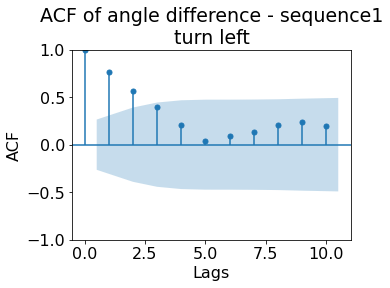}}
    \subfigure[]{\includegraphics[width=0.22\textwidth]{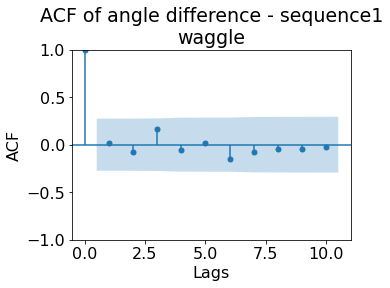}}
    \subfigure[]{\includegraphics[width=0.22\textwidth]{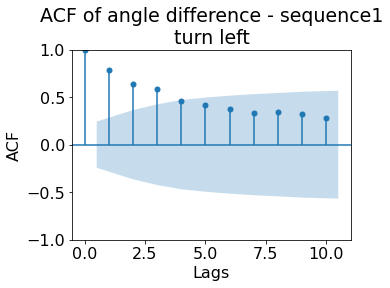}}
    \subfigure[]{\includegraphics[width=0.22\textwidth]{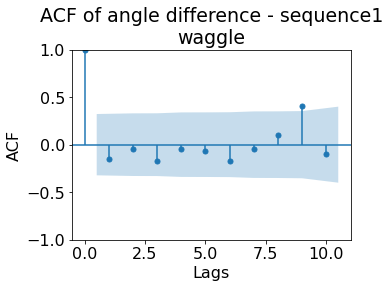}}
    \subfigure[]{\includegraphics[width=0.22\textwidth]{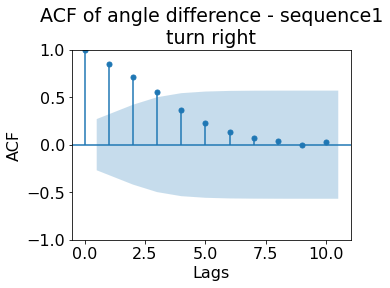}}
    \subfigure[]{\includegraphics[width=0.22\textwidth]{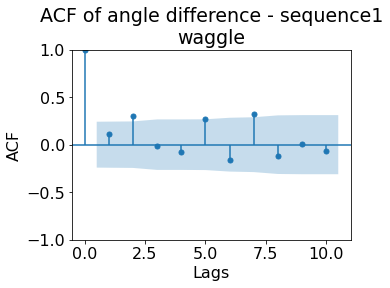}}  
    \subfigure[]{\includegraphics[width=0.22\textwidth]{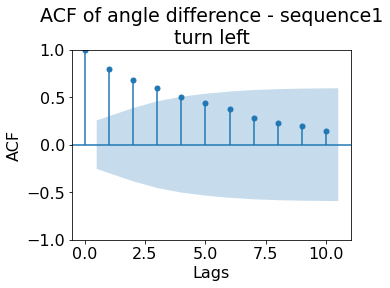}}
    \subfigure[]{\includegraphics[width=0.2\textwidth]{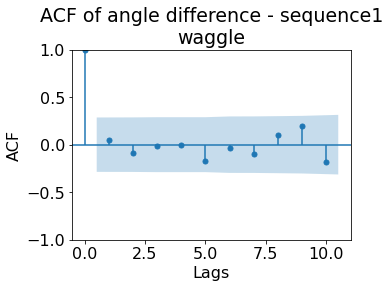}}
    \subfigure[]{\includegraphics[width=0.22\textwidth]{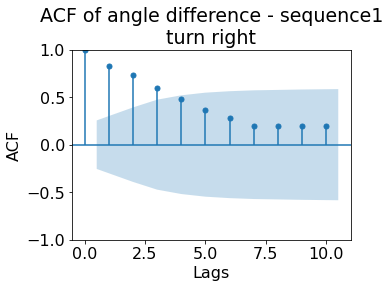}}
    \subfigure[]{\includegraphics[width=0.22\textwidth]{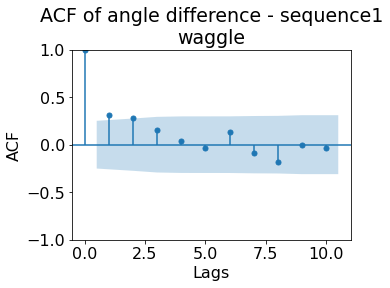}}
    \subfigure[]{\includegraphics[width=0.22\textwidth]{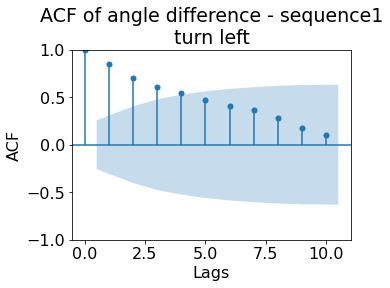}}
    \subfigure[]{\includegraphics[width=0.22\textwidth]{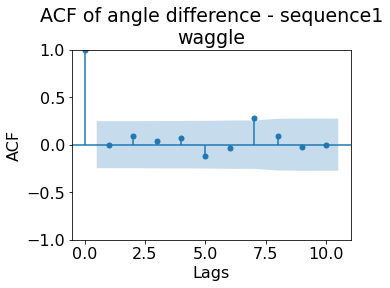}}
     \subfigure[]{\includegraphics[width=0.22\textwidth]{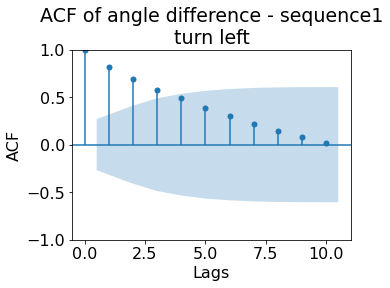}}
    \subfigure[]{\includegraphics[width=0.22\textwidth]{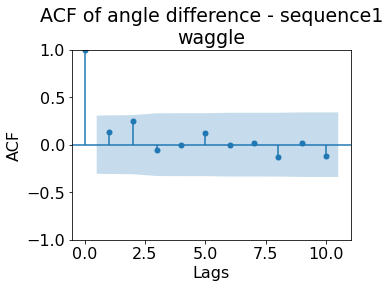}}
    \subfigure[]{\includegraphics[width=0.22\textwidth]{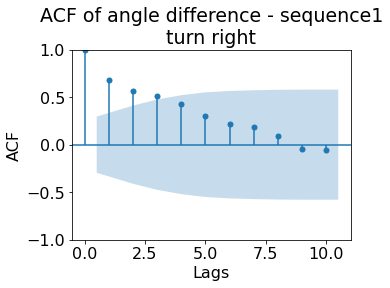}}
    \subfigure[]{\includegraphics[width=0.22\textwidth]{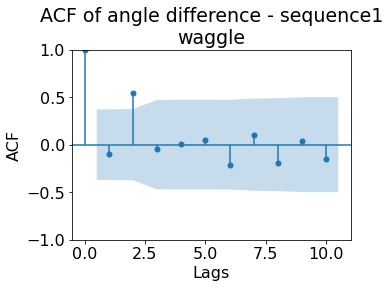}}
    \caption{The autocorrelation plot for the angle difference of the bee from sequence 1 in each ground true regime.}
    \label{fig: acf bees - true regimes}
\end{figure}

In this empirical analysis, we investigate the publicly available dancing bees dataset \cite{DancingBeesDataset}, which consists of trajectories $z_t = (x_t,y_t,\cos(\theta_t),\sin(\theta_t))$ and represents the location in 2D plane ($x_t$ and $y_t$) and the heading angle $\theta_t$ at time $t$ of the tracked bee, for six honey bees while performing the waggle dance. The dataset consists of six videos of sequences of the waggle dance. Each frame of the videos measures the position and the orientation of the bee over time. The waggle dance is one of the most popular examples of animal communication and is performed within the hive in order to communicate to the other bees the location and distance to a food source. The dance consists of three regimes; \textit{waggle, turn left, turn right} and is of great interest to biologists to identify the CPs between regimes. 

This example is important because (i) the true regimes are known, so we can assess in an objective way the detection performance of the different models; (ii) the angle difference variable is heteroscedastic in each true regime and autocorrelated in the majority of the true regimes, making it a perfect testbed for our Score-Driven models which are able to handle the corresponding variability of parameters. 

In order to infer the most likely CPs we work with the time series $\cos\theta_t-\sin\theta_t$ of the heading angle $\theta_t$ (top panel of Figure \ref{fig: MBOC_seq1}) for the first out of the six sequences, which as it seems from the autocorrelation plot within each true regime (see Figure \ref{fig: acf bees - true regimes}) it is a highly correlated time series. The rest of the dataset consists of the location of the bee in the 2D plane, which is depicted in the bottom panel of figure \ref{fig: MBOC_seq1} along with the ground truth regimes. 

The time series of the angle $\theta_t$ but also of the differences $\sin(\theta_t)-\cos(\theta_t)$ appears heteroskedasticity unconditionally and conditionally to the true regimes. We perform a Breusch-Pagan test on the entire dataset and on each regime separately, and we reject the null hypothesis of homoskedasticity for all the regimes and for the entire dataset at a $1\%$ significance level. Thus, we expect the MBOV model to capture this heteroskedasticity of the dataset and infer the CPs more accurately with respect to the other models. Moreover, we perform a Jarque-Bera (JB) test on the aforementioned datasets. The JB test rejects the null hypothesis of Gaussianity at the $5\%$ significance level for the entire data set, while it does not reject it at the $5\%$ significance level for $64\%$ of the regimes identified by the MBOV.

\begin{figure}[!h]
    \centering
    \subfigure[]{\includegraphics[width=0.24\textwidth]{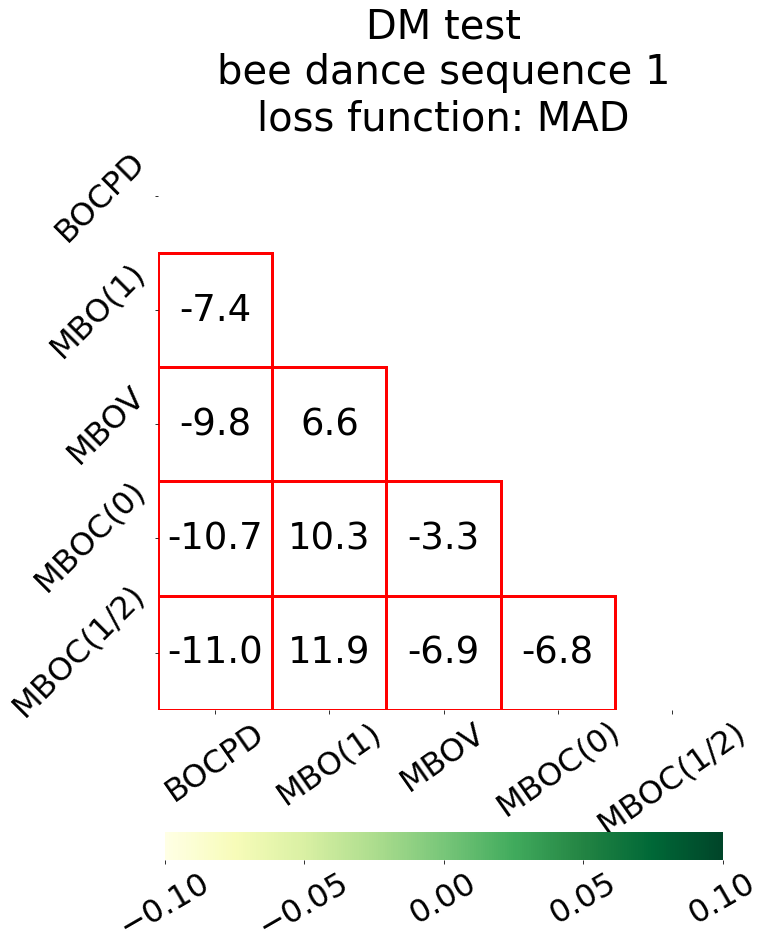}}  
    \subfigure[]{\includegraphics[width=0.24\textwidth]{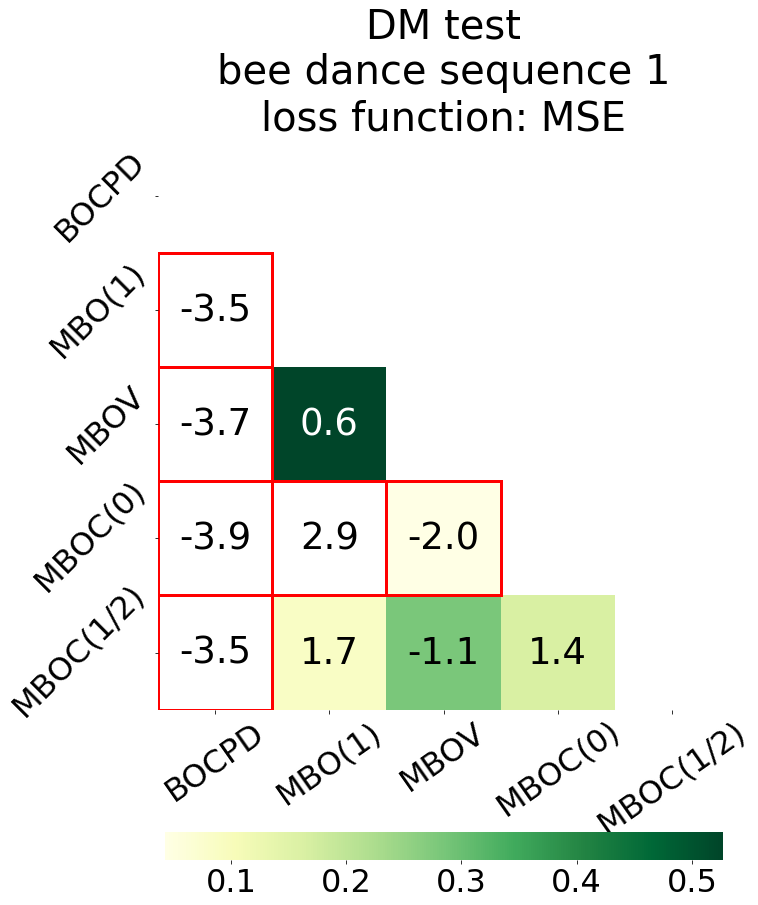}}
    \caption{Illustration of the Diebold-Mariano test for the sequence 1 (a) when the loss is the MAD in Eq. (\ref{eq: mad}) (b) when the loss is the MSE (see Eq.(\ref{eq: poly}) for $p=2$). The color of each square represents the p-value (the lighter the color the smaller the p-value). The white squares indicate a p-value less than $1\%$. The red squares indicate the significance at a $5\%$ level. The values within each square indicate the value of the DM statistic.}
    \label{fig: DM test Bees}
\end{figure}

Table \ref{tab: results on bees - seq1} reports the results for all the models for  one step ahead forecasting in terms of the MSE and the CM for the most likely regimes identified by each model. We observe that the best forecasting performance is obtained by MBOC(0) while the best identifiability of the regimes in terms of the CM is obtained by MBOV. Notice also that the second best performance is achieved by MBOC(1/2), indicating the importance of having time-varying parameters in the modeling.  The differences reported in table in the forecasting accuracy between BOCPD and the rest of the models are significant as it is illustrated in figure \ref{fig: DM test Bees} when it is performed the DM test, which rejects the null hypothesis of equal predictive accuracy at a $1\%$ significance level of the MBO(1), MBOV, MBOC(0) and MBOC(1/2) versus the BOCPD with both loss functions. The middle panel of Figure \ref{fig: MBOC_seq1} displays the run length posterior of MBOV for our dataset as well as the most likely path. We observe that the MBOV model captures well the majority of the ground true regimes.

For all the models, the initial mean value is set to $\mu_0 = 0$, the initial variance value to $\sigma^2_0=0.
3$ and the hazard rate to $h=1/80$. The known (initial) variance for the BOCPD (the rest of the models) is set to $\sigma^2=0.3$. The initial correlation for MBO(1), MBOV, MBOC($0$) and MBOC($1/2$) is set to $\rho=0.8$. The initial parameter vector for the MBOC($0$) and the MBOC($1/2$) is set to $\lambda=[0, 0.01, 0.9, 0.3]$.

\begin{center}
\begin{tabular}{c||c c c | c c c}
\hline
 & \multicolumn{3}{c|}{MSE}& \multicolumn{3}{c}{CM}\\
  $\rho$ & 0.7 & 0.8 & 0.9 & 0.7 & 0.8 & 0.9\\
\hline\hline
BOCPD & 0.141 & 0.141 & 0.141 & 0.606 & 0.606 & 0.606\\

MBO(1) & 0.126 & 0.120 & 0.117 & 0.561 & 0.595 & 0.559\\

\textbf{MBOV} & 0.119 & 0.119 & 0.119 & \textbf{0.617} & \textbf{0.617} & \textbf{0.617} \\ 

\textbf{MBOC(0)} & \textbf{0.116} & \textbf{0.116} & \textbf{0.116} & 0.595 & 0.595 & 0.595 \\  

MBOC(1/2) & 0.117 & 0.117 & 0.117 & 0.611 & 0.611 & 0.611\\
  \hline
\end{tabular}
\captionof{table}{Comparison among BOCPD, MBO($1$) MBOV, MBOC($0$) and MBOC($1/2$) of out of sample one-step-ahead MSE (columns 2-4) and of the CM (columns 5-7). The dataset refers to the bee dance of the first sequence. The correlation $\rho$ is the one used in MBO(1).}
\label{tab: Results for Waggle Bee Dance}
\label{tab: results on bees - seq1}
\end{center}

%===================================

\subsection{Growth Rate of Real GNP}
As we noted at the beginning of Section \ref{sec:methods}, a key benefit of the BOCPD methods is the avoidance of the prior imposition of a finite number of states for the mean. In this section we highlight this advantage by comparing the MBO(1) model with the Hamilton's Switching (HS) model in \cite{Hamilton1989}. Similarly to his work, here we use the growth rate of real U.S. GNP\footnote{The real GNP refers to the adjusted to inflation GNP. The (real) U.S. GNP dataset can be downloaded from \url{https://fred.stlouisfed.org/series/GNP}.} as the dataset for comparing the two models.

\begin{figure}[t]
\centering
\includegraphics[width=1\columnwidth]{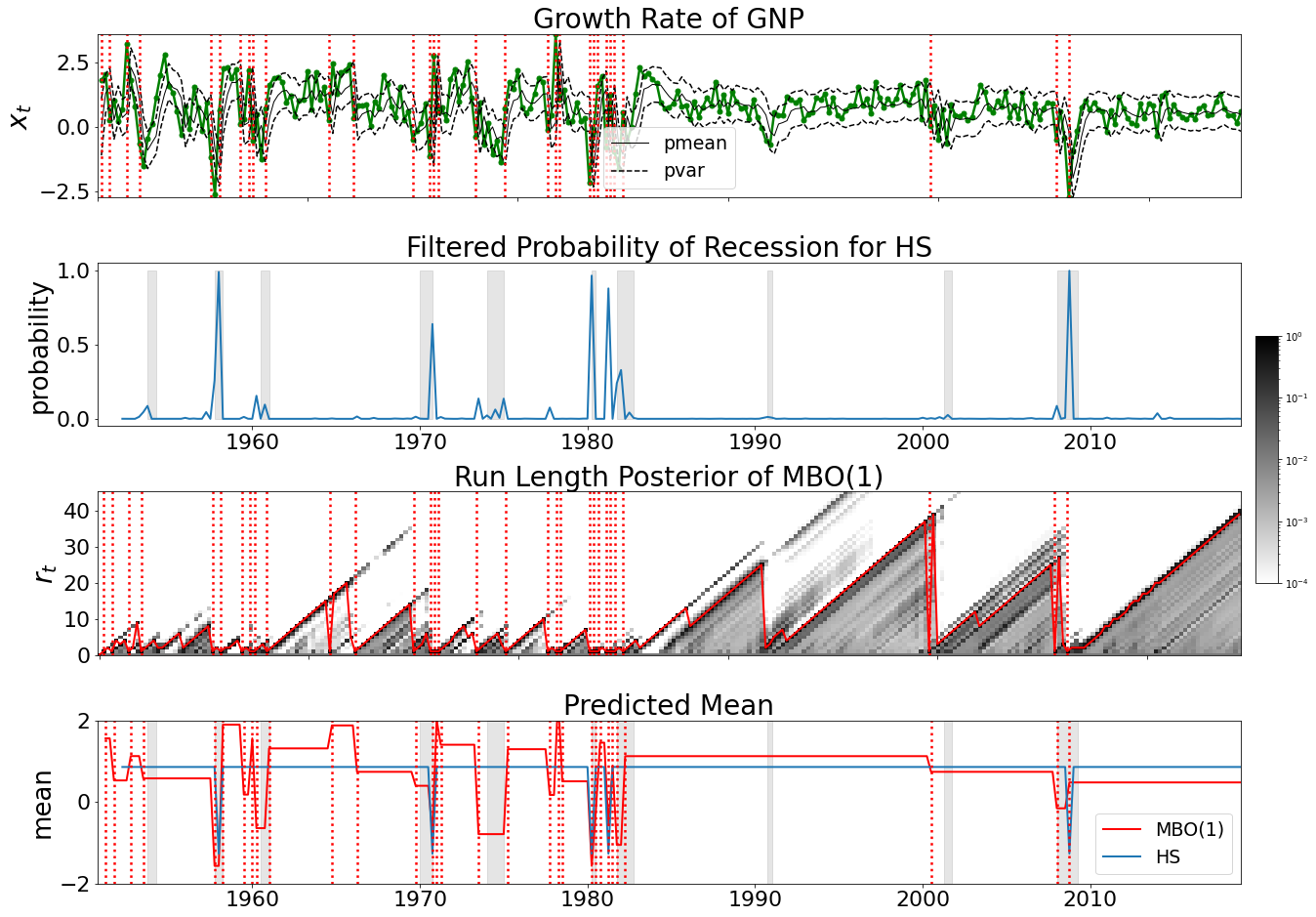}
\caption{\textbf{First:} The growth rate of U.S. GNP (green line) during $1951-2019$. The red dotted lines represent the CPs found by MBO(1). \textbf{Second:} The filtered probability of recession for HS model. The gray areas represent the periods of recession. The tick labels on the x-axis represent the first quarter of the indicated year. \textbf{Third:} The Run Length Posterior of MBO(1). The darkest the color, the highest the probability of the run length. The red solid line represents the most likely path i.e. value of $r_t$ with the largest run length posterior $p(r_t|x_{1:t})$ for each $t$. \textbf{Fourth:} The posterior mean of MBO(1) versus the predicted mean of HS model across the regimes identified by each method.}
\label{fig: GNP_MBO1_HS_V3}
\end{figure}

In this section we denote by $y_{1:T}$ the real U.S. GNP data which is quarterly  measured and spans in 1951-2019. In order to stabilize the variance of the dataset,  Hamilton considered the consecutive differences of the logarithm of GNP multiplied by 100, thus obtaining the growth rate of real GNP which is formally defined as:
\begin{equation}\label{eq: growth rate}
    x_t = 100\log\frac{y_t}{y_{t-1}}\quad\text{for}\quad t=2,\cdots,T.
\end{equation}
See the top panel of Figure \ref{fig: GNP_MBO1_HS_V3} for an illustration of $x_{2:T}$ (green plot). 

The HS model that is used for the study of real GNP is a 4-order autoregressive regime switching model with two states. The aim is to infer the slow and fast growth rates for the U.S. economy thus by assuming that there are two states, $s_t=0$ and $1$, associated with the recession and the growth period, respectively. The HS model operates offline and the autoregressive model's parameters are learned through the MLE using the complete dataset. Consequently, this approach is not suitable for real-time applications and can be highly sensitive to outliers. Notably, when the dataset contains periods of extreme events, such as the 2020 pandemic, the HS model struggles to detect recession periods. An extended analysis of the dependence of the two methods to the investigated period and to the presence of extreme events is presented in Appendix \ref{app:emp}.

The second panel of Figure \ref{fig: GNP_MBO1_HS_V3} shows the filtered probability of a recession for the HS model, i.e. $p(s_t=0|x_{1:t})$.
The third panel of Figure \ref{fig: GNP_MBO1_HS_V3} shows the run length posterior of MBO(1), using a gray scale shading to display, on the vertical axis,  the probability distribution of the run length (darker shades indicate higher probabilities). The red solid line traces the most likely path, representing the run length value $r_t$ with the highest posterior probability $p(r_t|x_{1:t})$ at each time $t$.

According to the two state HS model, the mean ($\mu_{s_t}$) is a function of the state and is found that
$$
\mu_{s_t} = \begin{cases}
    -1.267,\quad s_t=0\\
    0.856,\quad \quad s_t=1
\end{cases}.
$$
On the other hand, with MBO(1) model, which does not impose a specific number of states, we found that the mean ranges in a much wider spectrum than the one assumed by the HS model, with only two states. The bottom panel of Fig. \ref{fig: GNP_MBO1_HS_V3} compares the regimes and their mean value according to the two methods. For the HS model we choose a threshold value equal to $0.6$ and we plot the mean $\mu_{0}$ for every $t$ such that $p(s_t=0|x_{1:t})\geq 0.6$, otherwise we plot $\mu_1$. For the MBO(1) model we plot the posterior mean $\mu_t$ (see Eq. (\ref{eq: post params MBO1})) in each regime, found at the end of each regime. For the MBO($1$) we set as the prior mean value the $\mu_0=0$ and for the prior variance value the $\sigma_0^2 = 0.1$ and as correlation the value $\rho=0.1$. 

Next, we verify whether assumption (A2) holds for this dataset. Although the JB test rejects the null hypothesis of unconditional Gaussianity for U.S. real GNP, it fails to reject the null hypothesis at $5\%$ confidence level in $94\%$ of the regimes identified by MBO(1).

\subsection{Order Flow}
In this section we apply the BOCPD models to the time-series of aggregated order flow, which is the volume (i.e. number of shares) of buy market orders (i.e. orders that initiate a transaction) minus the
volume of sell market orders on a specific number of trades. The study of order flow is crucial in finance (see \cite{Lillo23}) because it involves the submission and execution of orders, which aggregate information and result in the formation of prices as orders arrive. In this application we perform one step ahead prediction of the order flow and we compare the significance of the predictions with the DM test. The order flow time series is known to exhibit long-range correlation, as documented by \cite{em:09}. Additionally, from an econometric perspective, this long-range correlation may relate to the established understanding (see \cite{r:01,MikoschStarica2004}) that long-memory time series can be approximately generated by regime-shift models. Therefore, models that account for autocorrelated data and regime-shifts are expected to perform better.

This application has been investigated by us in \cite{QFpaper}. The main novelty is the inclusion of models not considered there and the application of the DM test to show that the proposed models outperform in prediction the competitors in a statistically robust way. Our data set consists of orders during March 2020 for Microsoft Corp. (MSFT) and of December 2021 for Tesla Inc. (TSLA), provided by LOBSTER\footnote{LOBSTER (https://lobsterdata.com) provides high-frequency limit order book (LOB) data from NASDAQ. LOBs are electronic platforms through which the majority of stock trading occurs across most exchanges.}. Let $M$ represent the number of trades in a given day, and let $v_i$ ($i=1,\ldots,M$) denote the signed volume of the $i$-th trade (positive for buyer-initiated trades and negative for seller-initiated trades). We aggregate $N$ trades to form our time series, resulting in $T=\lfloor M/N \rfloor$ observations. The time series of interest $x_t$, represents the aggregated order flow over the interval $~\N\cap[N(t-1)+1,N(t-1)+N]$ and is given by 
\begin{equation}
    x_t = \sum_{j=1}^N v_{N(t-1)+j},~~~~~~~~~~t=1,...,T.
\end{equation}
For TSLA we set $N=970$ and for MSFT $N=1620$ executions which, in both cases correspond to an average time interval of 4 minutes. The length of the two time series is 2150 data points for TSLA and 2154 for MSFT.

We make one step ahead forecast by computing at each time step $t$ the predictive mean $\hat{\mu}_t$ (see Eq.(\ref{eq: pred mean})).
Table \ref{tab: Results Order Flow} shows the results of the comparison among the various models in terms of the MSE (see Eq.(\ref{eq: MSE})). From the Table it is apparent that for Tesla when $\Delta t=4$min the best forecasting performance is by MBOC(0), hence from the MBOC model when $d=0$ in the Fisher information matrix. Very similar performance it is also observed from MBOC when $d=1/2$ and from MBOV models. The ARMA(1,1) model, that does not infer the position of the CPs displays similar performance with the BOCPD model which does not account for a Markovian correlation within regimes. All the above results when $\Delta t=4$min, are coherent with different time scales that have been considered in a previous study in \cite{QFpaper} along with an analysis on the predictability of the identified regimes by MBOC(0) in the order flow. 

To apply the models effectively, it is important to carefully select several hyperparameters. For all models, and both stocks the initial mean value is specified as $\mu_0 = 0$ and the hazard rate as $h=1/80$. For TSLA the known variance value for the BOCPD is set to $\sigma_0^2=4\times 10^7$ while for MSFT to $\sigma_0^2=7\times 10^8$. For the rest of the models that require an initial variance in order to start the inference, we use the same value as it has been used for the BOCPD. The initial correlation for the Score-Driven MBO(1) is set to $\rho=0.2$ and $0.3$ for TSLA and MSFT respectively. The initial parameter vector for MBOC($0$) and MBOC($1/2$) is set to $\vec\lambda = [0, 0.01, 0.9, 10^7]$ for both stocks.

\begin{figure}[!h]
    \centering
    \subfigure[]{\includegraphics[width=0.24\textwidth]{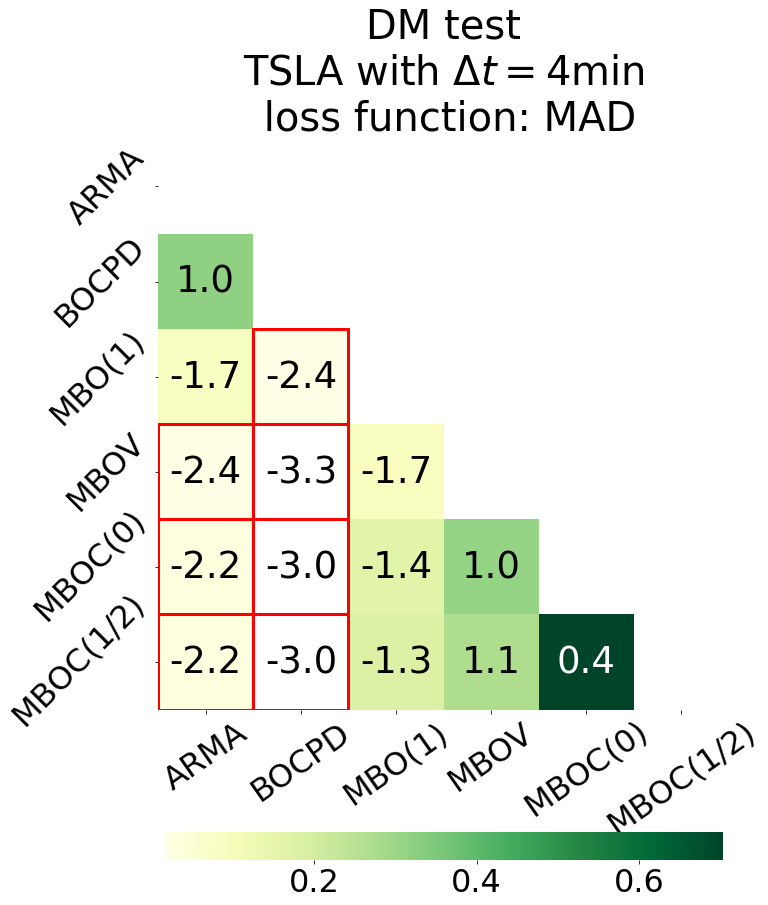}}  
    \subfigure[]{\includegraphics[width=0.24\textwidth]{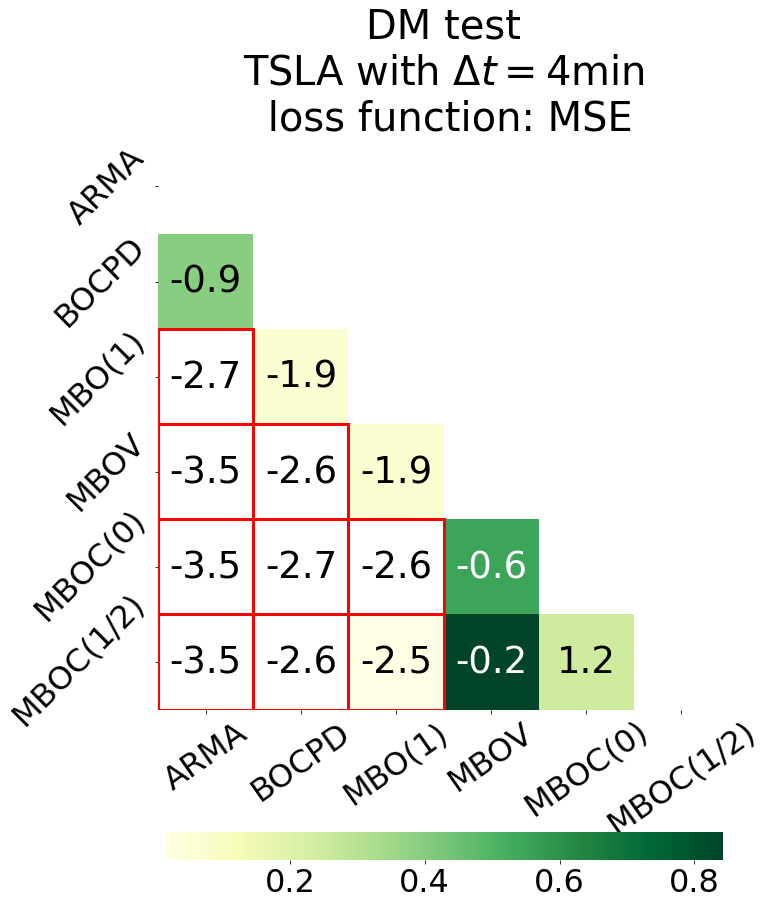}}
    \subfigure[]{\includegraphics[width=0.24\textwidth]{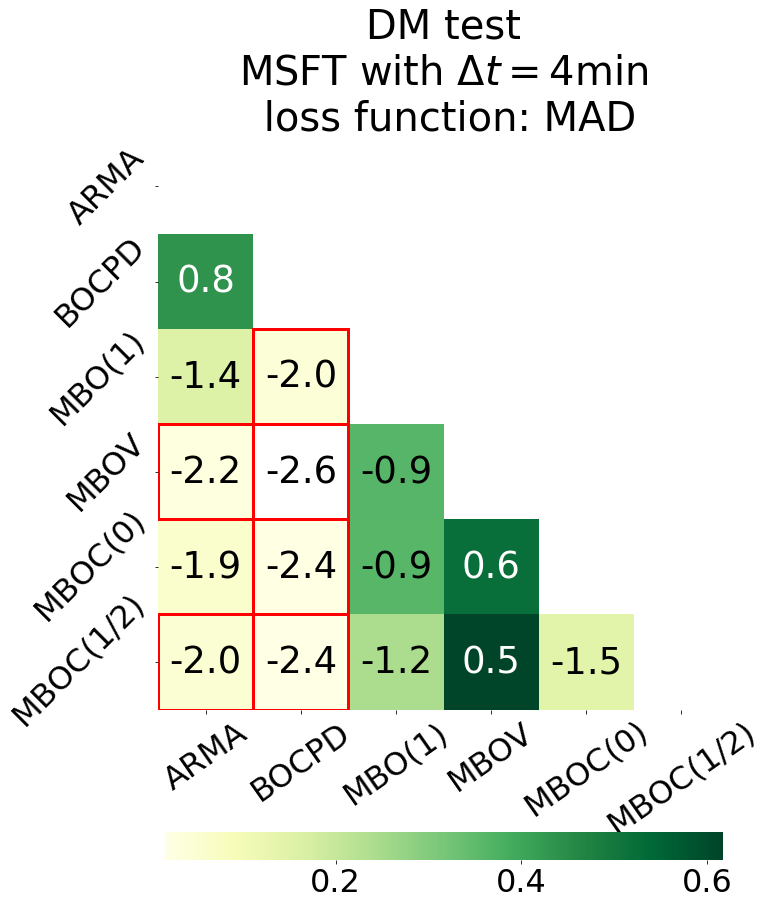}}  
    \subfigure[]{\includegraphics[width=0.24\textwidth]{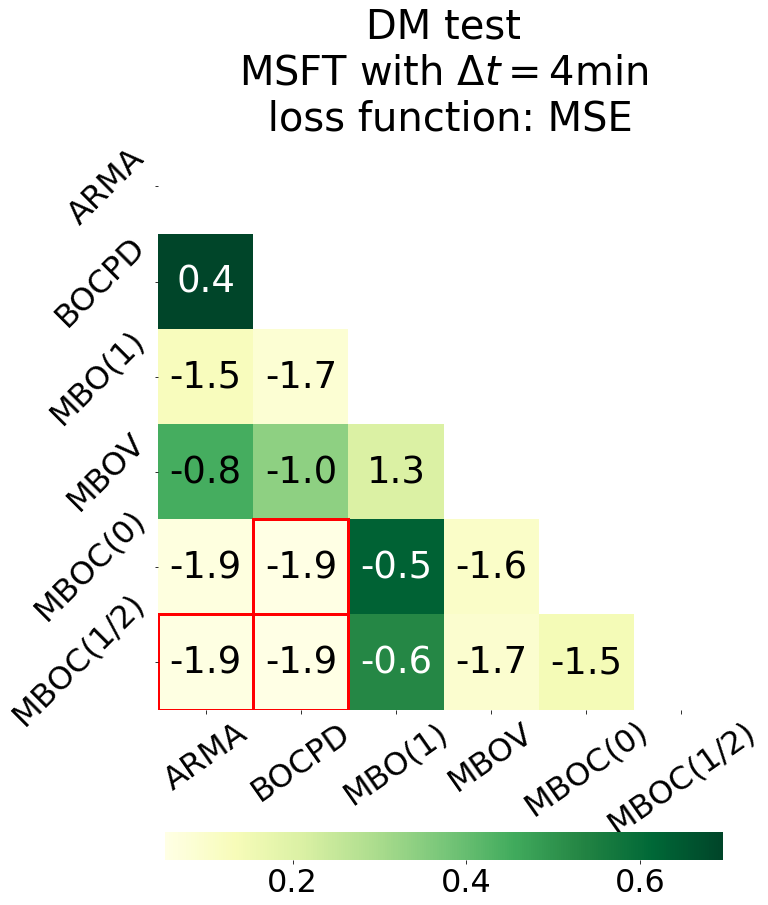}}
    \caption{Illustration of the DM test for TSLA with $\Delta t=4$min (a) when the loss is the MAD in Eq. (\ref{eq: mad}) (b) when the loss is the MSE (see Eq.(\ref{eq: poly}) for $p=2$) and for MSFT with $\Delta t=4$min (c) when the loss is the MAD (d) when the loss is the MSE. The color of each square represents the p-value (the lighter the color the smaller the p-value). The white squares indicate a p-value less than $1\%$. The red squares indicate the significance at $5\%$ level. The values within each square indicate the value of the DM statistic. A negative value indicates that the model in the row has better forecasting power than the one in the column.}
    \label{fig: DM test Order Flow}
\end{figure}

In order to test the statistical significance of these differences in the predictive performance we use the DM test. The residuals for the Bayesian class models (BOCPD, MBOV, MBOC(0), and MBOC(1/2)) are defined as $e_i = \hat{\mu}_i - x_i$, where $\hat{\mu}_i$ is the predictive mean at time $i$.
The results of the DM test are illustrated in  Figure \ref{fig: DM test Order Flow}. The color in each rectangle indicates the p-value, the lighter the color the smaller the p-value, while the reported number in each cell indicates the DM statistic. In particular, the DM test for TSLA with $\Delta t=4$min rejects the null hypothesis of equal predictive accuracy at a $1\%$ significance level for the MBOV, MBOC(0) and MBOC(1/2) versus the ARMA(1,1) and the BOCPD for both loss functions. For MSFT with $\Delta t=4$min it rejects the null hypothesis for the MBOC(0) and MBOC(1/2) versus the ARMA(1,1) and the BOCPD model at a $5\%$ significance level for both loss functions. The DM test, relies on a unique assumption, that of stationarity of the loss-differential time series which is satisfied since the Augmented Dickey-Fuller test we performed in the five loss-differential time series for each stock, rejects the null hypothesis of non-stationarity at a $1\%$ significance level.

Finally, we test that assumption (A2) is satisfied. Specifically, even though the JB test rejects the Gaussianity null hypothesis of the unconditional aggregated order flow at $5\%$ significance level for both stocks, it does not reject the null hypothesis at $5\%$ significance level for $95\%$ of the regimes found by MBOC(0). These results are in line with the results in \cite{QFpaper} when different time scales $\Delta t$ are also considered.

\begin{center}
\begin{tabular}{c||c c c|c c c}
\hline
 & \multicolumn{3}{c|}{TSLA} & \multicolumn{3}{c}{MSFT} \\
\cline{1-7}
  $\rho$ & 0.1 & 0.2 & 0.3 & 0.2 & 0.3 & 0.4\\
\hline\hline
ARMA & 0.896 & 0.896 & 0.896 & 0.844  & 0.844 & 0.844\\

BOCPD & 0.888 & 0.888 & 0.888 & 0.847 & 0.847 & 0.847\\

MBO(1) & 0.884 & 0.875 & 0.880 & 0.843 & 0.833 & 0.838\\ 

MBOV & 0.872 & 0.872 & 0.872 & 0.833 & 0.833 & 0.833\\
  
   \textbf{MBOC(0)} & \textbf{0.870} & \textbf{0.870} & \textbf{0.870} & \textbf{0.832} & \textbf{0.832} & \textbf{0.832}\\
\textbf{MBOC(1/2)} & 0.871 & 0.871  & 0.871  & \textbf{0.832} & \textbf{0.832} & \textbf{0.832}\\
  \hline
\end{tabular}
\captionof{table}{Comparison of out of sample one-step-ahead MSE of ARMA(1,1), BOCPD, MBOV, MBOC(0) and MBOC(1/2) when $\Delta t=4$min.}
\label{tab: Results Order Flow}
\end{center}

%================================

\section{Conclusion}\label{sec:concl}
In this study, we present a new methodology for CPD in univariate time series with different regimes in the mean. We build upon the BOCPD model introduced by \cite{c:07}, which detects CPs in real time without predefining a finite number of states. A key limitation of BOCPD is its assumption of independent data within each regime's DGP. We address this limitation by incorporating temporal dependence of observations with any memory order $q\in\N^*$ within regimes and deriving closed-form solutions for the UPM for any $q\in\N^*$. Introducing autoregressive dynamics within regimes is crucial, as the independence assumption is rarely met in real-world applications. Furthermore, by integrating the Score-Driven AR(1) process introduced by \cite{Score-Driven_AR}, we account for time-varying parameters (other than the mean) that capture dynamic patterns in the data. This extension makes our methodology particularly useful for finance-related problems, where time-varying parameters such as variance or correlation are common.

We demonstrate the effectiveness of our methodology by applying it first to simulated data and then to real data sets. In the simulations, our proposed autoregressive extension MBO(1) outperforms the baseline model BOCPD in terms of forecasting and CPD when the data is autocorrelated, even with misspecified parameter dynamics. We also highlight the importance of detecting CPs for the Score-Driven AR(1) model to accurately recover the true dynamics of autocorrelation. For the empirical analysis, we use three different data sets, each with unique characteristics. First, we analyze the so-called bee waggle dance data set,  which exhibits both autocorrelation and heteroskedasticity within each regime. Here, the Score-Driven MBO(1) model shows the best forecasting accuracy at the $1\%$ significance level compared to other models, while the MBOV model performs best in regime identification based on the covering metric.  Second, we compare the MBO(1) model with the Markov Switching model introduced by \cite{Hamilton1989} using real U.S. GNP data to identify the recession periods of the U.S. economy. The MBO(1) model successfully identifies these periods and due to its flexibility in not specifying the number of states, also distinguishes between significant and less significant recession or growth periods. Finally, we analyze the order flow data set for two liquid stocks traded in the NASDAQ stock market, which is highly correlated for both stocks and observe that the Score-Driven MBO(1) significantly outperform both the baseline and the ARMA model, which assumes autoregressive data but does not identify regimes, in terms of forecasting accuracy. These differences are significant at the $5\%$ level according to a  DM test. 

Future developments of this work could include incorporating time-varying parameters other than the mean into the MBO($q$) model for $q$ greater than one. Specifically, this would involve the development of the Score-Driven MBO($q$) for $q>1$. Additionally, there is potential for extending the methodology to identify CPs in other parameters of the time series, such as variance or correlation, particularly when the data is dependent within regimes. Finally, extending the MBO($q$) model to multivariate time series would also be an interesting direction for further research.

\section*{Declarations}
\subsection*{Acknowledgement}
IYT acknowledges 
Daniele Maria Di Nosse and Edoardo Urettini for useful discussions. FL gratefully acknowledges financial support from SoBigData.it, which receives funding from European Union – NextGenerationEU – PNRR – Project: “SoBigData.it – Strengthening the Italian
RI for Social Mining and Big Data Analytics” – Prot. IR0000013 – Avviso n. 3264 del 28/12/2021. PM and IYT acknowledge financial support by the Italian Ministry of University and Research under the PRIN project {\it Realized Random Graphs: A New Econometric Methodology for the Inference of Dynamic Networks} (grant agreement n. 2022MRSYB7, CUP B53D23010100001).

%PM acknowledges financial support under the National Recovery and Resilience Plan (NRRP), Mission 4, Component 2, Investment 1.1, Call for tender No. 1409 published on 14.9.2022 by the Italian Ministry of University and Research (MUR), funded by the European Union – NextGenerationEU– Project Title ``Climate Change, Uncertainty and Financial Risk: Robust Approaches based on Time-Varying Parameters'' - CUP B53D23026390001 - Grant Assignment Decree No. P20229CJRS adopted on 27.11.2022 by the Italian Ministry of University and Research (MUR).%financial support under the National Recovery and Resilience Plan (NRRP), Mission 4, Component 2, Investment 1.1, Call for tender No. 104 published on 2.2.2022 by the Italian Ministry of University and Research (MUR), funded by the European Union – NextGenerationEU– Project Title “Realized Random Graphs: A New Econometric Methodology for the Inference of Dynamic Networks” – CUP B53D23010100001 - Grant Assignment Decree No. 2022MRSYB7 by the Italian Ministry of Ministry of University and Research (MUR).

\appendix

\section{MBO($q$) Derivations}\label{app: MBO(q) Derivations}
In this section we will prove proposition~(\ref{prop: MBO(q)}). Let $x_t^{(r_t)}$ be the data in regime $R$ and $j^*=\min\{q,i-1-t+r_t\}$ then we will show that $p(\theta_R) = \mathcal{N}(\mu_0,\sigma_0^2)$, is a conjugate prior, hence the posterior mean is also normally distributed. In the following we will use notation 
\begin{equation}\label{eq: dependent mu and sigma}
\mu_{x_i|x_{i-1}^{(j^*)}} = \theta_R+\begin{bmatrix}
        \gamma_1\\
        \vdots\\
        \gamma_{j^*}
\end{bmatrix}^\intercal\Sigma_{j^*}^{-1}\begin{bmatrix}
    x_{i-1}-\theta_R\\
    \vdots\\
    x_{i-j^*}-\theta_R
\end{bmatrix}\quad\text{and}\quad
    v_{j^*} =  \gamma_0-\begin{bmatrix}
        \gamma_1\\
        \vdots\\
        \gamma_{j^*}
\end{bmatrix}^\intercal\Sigma_{j^*}^{-1}\begin{bmatrix}
        \gamma_1\\
        \vdots\\
        \gamma_{j^*}
    \end{bmatrix}
\end{equation}
for the conditional mean and variance respectively of $x_i$ given $x_{i-1}^{(j^*)}$. Since for every $x_i\in x_t^{(r_t)}$ the assumption (A2) is satisfied we can write the conditional probability distribution:
\begin{align}
    p(x_i|x_{i-1}^{(j^*)}) & = \frac{1}{\sqrt{2\pi v_{j^*}}}\exp\bigg\{-\frac{1}{2}\frac{\Big(x_i-\mu_{x_i|x_{i-1}^{(j^*)}}\Big)^2}{v_{j^*}}\bigg\}\nonumber\\
    & \propto\exp\Big\{-\frac{1}{2v_{j^*}}(\underbrace{\begin{bmatrix}
x_{i}-\theta_R\\
x_{i-1}-\theta_R\\
\vdots\\
x_{i-j^*}-\theta_R
\end{bmatrix}^\intercal}_{y_{j^*+1}^\intercal}\underbrace{\begin{bmatrix}
    1\\
    -\Sigma_{j^*}^{-1}\begin{bmatrix}
        \gamma_1\\
        \vdots \\
        \gamma_{j^*}    
\end{bmatrix}
\end{bmatrix}}_{c_{j^*+1}})^2\Big\}.
\end{align}
Let the matrix $C_{j^*+1} = c_{j^*+1}c_{j^*+1}^\intercal$. 
% which can also be written as
% $$
% C_{j^*+1} = \frac{1}{d_{j^*}^2}\begin{pmatrix}
%     d_{j^*}^2 & -d_{j^*}\sum_{i=1}^{{j^*}} (-1)^{i-1}\gamma_{i}\gamma_{i-1} & \cdots & -d_{j^*}\sum_{i=1}^{{j^*}} (-1)^{i-{j^*}}\gamma_{i}\gamma_{i-{j^*}}\\
%     -d_{j^*}\sum_{i=1}^{{j^*}} (-1)^{i-1}\gamma_{i}\gamma_{i-1} & \Big(\sum_{i=1}^{{j^*}} (-1)^{i-1}\gamma_{i}\gamma_{i-1}\Big)^2 & \cdots & \sum_{i=1}^{{j^*}}\sum_{k=1}^{{j^*}} (-1)^{i+k-1-{j^*}}\gamma_{i}\gamma_{i-1}\gamma_{k}\gamma_{k-{j^*}}\\
%     \vdots & \vdots & \cdots & \vdots\\
%     -d_{j^*}\sum_{i=1}^{{j^*}} (-1)^{i-{j^*}}\gamma_{i}\gamma_{i-{j^*}} & \cdots & \cdots & \Big(\sum_{i=1}^{{j^*}} (-1)^{i-{j^*}}\gamma_{i}\gamma_{i-{j^*}}\Big)^2
% \end{pmatrix}
% $$
% with $d_{j^*}$ we denote the determinant of matrix $\Sigma_{j^*}$. 
By decomposing $y^\intercal_{j^*+1} = \pmb{x}_{i}^{(j^*+1)}-\pmb{\theta}_R^{j^*+1}$ with $\pmb{x}_{i}^{(j^*+1)}=\begin{bmatrix}
    x_i\\
    \vdots\\
    x_{i-j^*}
\end{bmatrix}$ and $\pmb{\theta}_R^{j^*+1} = \begin{bmatrix}
    \theta_R\\
    \vdots\\
    \theta_R
\end{bmatrix}$ a vector of dimension $j^*+1$, the conditional distribution can be written as
\begin{align}
    p(x_i|x_{i-1}^{(j^*)})& \propto\exp\Big\{-\frac{1}{2v_{j^*}}(\pmb{x}_{i}^{(j^*+1)}-\pmb{\theta}_R^{j^*+1})^\intercal C_{j^*+1}(\pmb{x}_{i}^{(j^*+1)}-\pmb{\theta}_R^{j^*+1})\Big\}\nonumber\\
   & \propto\exp\Big\{-\frac{1}{v_{j^*}} \Big( -2(\pmb{x}_{i}^{(j^*+1)})^\intercal C_{j^*+1}\pmb{\theta}_R^{j^*+1} + (\pmb{\theta}^{j^*+1}_R)^\intercal C_{j^*+1}\pmb{\theta}_R^{j^*+1}\Big)\Big\}.
\end{align}
Then, to find the posterior hyperparameters we apply the Bayes' rule:
\begin{align}
    p(\theta_R|x_t^{(r_t)})& \propto p(x_t^{(r_t)}|\theta_R)p(\theta_R)\nonumber\\
     & = p(\theta_R)\prod_{i=t+1-r_t}^{t}p(x_i|x_{i-1}^{(j^*)})\quad\text{with}\quad j^*=\min\{q,i-1-t+r_t\}\nonumber\\
     & \propto \exp\Big\{-\frac{1}{2\sigma_0^2}(\theta_R^2-2\theta_R\mu_0)\Big\}\nonumber\\&\prod_{i=t+1-r_t}^{t}\exp\Big\{-\frac{1}{2v_{j^*}} \Big( -2(\pmb{x}_{i}^{(j^*+1)})^\intercal C_{j^*+1}\pmb{\theta}_R^{j^*+1} + (\pmb{\theta}^{j^*+1}_R)^\intercal C_{j^*+1}\pmb{\theta}_R^{j^*+1}\Big)\Big\}\nonumber\\\label{eq: likelihood mbo(q) 2}
     & = \exp\Big\{-\frac{1}{2}\Big[-2\Big(\sum_{i=q+1+t-r_t}^t\pmb{x}_{i}^{(q+1)}\Big)^\intercal\frac{C_{q+1}}{v_{q}}\pmb{\theta}_R^{q+1}+(\pmb{\theta}_R^{q+1})^\intercal\frac{(r_t-q)C_{q+1}}{v_{q}}\pmb{\theta}_R^{q+1}+{\theta}_R^2\frac{1}{\sigma_0^2}-2\frac{\mu_0}{\sigma_0^2}{\theta}_R\\\label{eq: likelihood mbo(q) 3}
    &+\sum_{i=t+1-r_t}^{q+t-r_t}\Big[-2\big( \pmb{x}_{i}^{(i-t+r_t)}\big)^\intercal \frac{C_{i-t+r_t}}{v_{i-1-t+r_t}}\pmb{\theta}_R^{i-t+r_t}+(\pmb{\theta}_R^{i-t+r_t})^\intercal\frac{C_{i-t+r_t}}{v_{i-1-t+r_t}}\pmb{\theta}_R^{i-t+r_t}\Big]\Big]\Big\}\\\label{eq: likelihood mbo(q) 4}
     & = \exp\Big\{-\frac{1}{2}\Big[(\pmb{\theta}_R^{s})^\intercal A_{r_t}\pmb{\theta}_R^{s}-2b_{r_t}^T\pmb{\theta}_R^{s}\Big]\Big\}\quad\text{where}\quad s= \frac{(q+1)(q+2)+2}{2}.
\end{align}
Notice that for any $i\geq q+1+t-r_t$ it is true that $i-1-t+r_t\geq q$ and thus $j^*=q$ hence line (\ref{eq: likelihood mbo(q) 2}) follows. On the other hand, if $i\leq q+t-r_t$  it is true that $i-1-t+r_t<q$ thus $j^*=i-1-t+r_t$ and line (\ref{eq: likelihood mbo(q) 3}) follows. The matrices $A_{r_t}$ and $b_{r_t}^\intercal$ are written as:
\begin{align}
    A_{r_t} & = \begin{pmatrix}
        \frac{(r_t-q)C_{q+1}}{v_{q}} & 0 & 0 & \cdots & 0 & 0 & 0\\
        0 & \frac{C_{q}}{v_{q-1}} & 0 & \cdots & 0 & 0 & 0\\
        0 & 0 & \frac{C_{q-1}}{v_{q-2}} & 0 & \cdots & 0 & 0\\
        0 & 0 & 0 & \frac{C_{q-2}}{v_{q-3}} & \cdots & 0 & 0\\
        \vdots & \vdots & \vdots & \cdots & \vdots & \vdots & \vdots\\
        0 & 0 & 0 & \cdots & 0 & \frac{C_1}{v_0} & 0\\
        0 & 0 & 0 & \cdots & 0 & 0 &\frac{1}{\sigma_0^2}
\end{pmatrix} \quad\text{and}\quad A_{r_t} = (a^{r_t}_{ij})_{1\leq i,j\leq s}\in\R^{s\times s}
\end{align}
and
\begin{align}
    b_{r_t}^T = \begin{pmatrix}
        \sum_{i=q+1+t+r_t}^t x_i\\
        \vdots\\
        \sum_{i=q+1+t+r_t}^t x_{i-q}\\
        \pmb{x}_{q+t-r_t}^{(q)}\\
        \vdots\\
        \pmb{x}_{1+t-r_t}^{(1)}\\
        1
    \end{pmatrix}^T \begin{pmatrix}
         \frac{C_{q+1}}{v_{q}} & 0 & 0 & \cdots & 0 & 0 & 0\\
         0 & \frac{C_{q}}{v_{q-1}} & 0 & \cdots & 0 & 0 & 0\\
        0 & 0 & \frac{C_{q-1}}{v_{q-2}} & 0 & \cdots & 0 & 0\\
        0 & 0 & 0 & \frac{C_{q-2}}{v_{q-3}} & \cdots & 0 & 0\\
        \vdots & \vdots & \vdots & \cdots & \vdots & \vdots & \vdots\\
        0 & 0 & 0 & \cdots & 0 & \frac{C_1}{v_0} & 0\\
        0 & 0 & 0 & \cdots & 0 & 0 & \frac{\mu_0}{\sigma_0^2}
    \end{pmatrix}\quad\text{and}\quad b_{r_t}^T = (b^{r_t}_{j})_{1\leq j\leq s}\in\R^{1\times s}.
\end{align}
We thus have
\begin{align}\label{eq: final form of posterior}
    p(\theta_R|x_{t}^{(r_t)}) & \propto \exp\Big\{-\frac{1}{2}\Big[(\pmb{\theta}_R^{s})^\intercal A_{r_t}\pmb{\theta}_R^{s}-2b_{r_t}^T\pmb{\theta}_R^{s}\Big]\Big\}
\end{align}
By expanding the RHS of Eq.~(\ref{eq: final form of posterior}) we get
    \begin{align}
    p(\theta_R|x_{t}^{(r_t)}) & \propto \exp\Big\{-\frac{1}{2}\Big[\theta_R^2\underbrace{\sum_{1\leq i,j\leq s}a^{r_t}_{ij}}_{a_{r_t}}-2\theta_R\underbrace{\sum_{1\leq j\leq s}b^{r_t}_j}_{b_{r_t}}]\Big\}\nonumber\\\label{eq: compl square}
    & \propto \exp\Big\{-\frac{a_{r_t}}{2}\Big(\theta_R-\frac{b_{r_t}}{a_{r_t}}\Big)^2\Big\}.
\end{align}
Line (\ref{eq: compl square}) is obtained by completing the square. The posterior parameters for the MBO($q$) model are:
\begin{equation}\label{eq: post params final}
    \sigma_{r_t}^2  = \frac{1}{a_{r_t}}\quad\text{and}\quad
    \mu_{r_t}  = \frac{b_{r_t}}{a_{r_t}}
\end{equation}
hence
\begin{equation}
    p(\theta_R|x_t^{(r_t)}) = \mathcal{N}(\mu_{r_t},\sigma_{r_t}^2)
\end{equation}
and the UPM:
\begin{align}
    p(x_{t+1}|x_{t}^{(r_t)}) & = \int p(x_{t+1}|x_t^{(r_t)},\theta_R)p(\theta_R|x_t^{(r_t)})d\theta_R\nonumber\\
    & = \int p(x_{t+1}|x_t^{(q)},\theta_R)p(\theta_R|x_t^{(r_t)})d\theta_R\nonumber\\
    & = \int \mathcal{N}(\theta_R+\begin{bmatrix}
        \gamma_1\\
        \vdots\\
        \gamma_q
    \end{bmatrix}^\intercal\Sigma_q^{-1}\begin{bmatrix}
        x_{t-1}-\theta_R\\
        \vdots\\
        x_{t-q}-\theta_R
    \end{bmatrix},\overbrace{\gamma_0-\begin{bmatrix}
        \gamma_1\\
        \vdots\\
        \gamma_q
    \end{bmatrix}^\intercal\Sigma_q^{-1}\begin{bmatrix}
        \gamma_1\\
        \vdots\\
        \gamma_q
    \end{bmatrix}}^{v_{q}})\mathcal{N}(\mu_{r_t},\sigma_{r_t}^2)d\theta_R\nonumber\\
    & =\int\mathcal{N}(\underbrace{\pmb{1}^{q+1}c_{q+1}}_{A}\underbrace{\theta_R}_{x}+\underbrace{\begin{bmatrix}
        \gamma_1\\
        \vdots\\
        \gamma_q
    \end{bmatrix}^\intercal\Sigma_q^{-1}\pmb{x}_{t-1}^{(q)}}_{b},\underbrace{\gamma_0-\begin{bmatrix}
        \gamma_1\\
        \vdots\\
        \gamma_q
    \end{bmatrix}^\intercal\Sigma_q^{-1}\begin{bmatrix}
        \gamma_1\\
        \vdots\\
        \gamma_q
    \end{bmatrix}}_{L^{-1}})\mathcal{N}(\underbrace{\mu_{r_t}}_{\mu},\underbrace{\sigma_{r_t}^2}_{\Lambda^{-1}})d\theta_R\nonumber\\
    &=\mathcal{N}(\pmb{1}^{q+1}c_{q+1}\mu_{r_t}+\begin{bmatrix}
        \gamma_1\\
        \vdots\\
        \gamma_q
    \end{bmatrix}^\intercal\Sigma_q^{-1}\pmb{x}_{t-1}^{(q)},v_q+\pmb{1}^{q+1}c_{q+1}\sigma_{r_t}^2(\pmb{1}^{q+1}c_{q+1})^\intercal).
\end{align}
The last equality is due to the Bayes' theorem:
\begin{align*}
    \text{If}\quad p(x) & = \mathcal{N}(\mu,\Lambda^{-1})\\
   \text{and}\quad p(y|x) & = \mathcal{N}(Ax+b,L^{-1})\\
   \text{then}\quad p(y) & = \mathcal{N}(A\mu+b,L^{-1}+A\Lambda^{-1}A^T).
\end{align*}
% With the following substitutions:
% \begin{align}
%     A & = \pmb{1}_{j^*+1}^Tc_{j^*+1}\\
%     x & = \mu\\
%     b & = \Sigma_{\{1\},\{1+j\}_{j=1}^{j^*}}\Sigma_{\{j\}_{j=1}^{j^*},\{j\}_{j=1}^{j^*}}^{-1}\{\pmb{x}_{n+1-j}\}_{j=1}^{j^*}\\
%     L^{-1} & = \Sigma_{\{n+1\}|\{n+1-j\}_{1\leq j\leq j^*}}\\
%     \mu & = \mu_n\\
%     \Lambda^{-1} & = \sigma_n^2
% \end{align}
% we get the result.

%==================================
\section{Connection with Maximum Likelihood}\label{app: Connection with Maximum Likelihood}
In this section we prove proposition \ref{prop1}.
\newline\newline
Let $x_t^{(r_t)}$ satisfying assumption (A2) for $q=1$ and consider $r_t=t$. The aim is to find the parameter $\theta_R$ through the Maximum Likelihood Estimation. Thus we consider the log-likelihood
% \begin{align}
%     \log p(x_{1:t}|\mu,\sigma^2) & = \log \prod_{i=1}^t p(x_i|\mu,\sigma^2)\nonumber\\
%     & = -\frac{1}{2\sigma^2}\sum_{i=1}^t (x_i-\mu)^2-\frac{t}{2}\log \sigma^2-\frac{t}{2}\log 2\pi 
% \end{align}
% If now the sample is dependent of order $q=1$ the log-likelihood is:
\begin{align}
    \log p(x_{1:t}|\theta_R,\gamma_0,\gamma_1) & = \log p(x_1)+\sum_{i=2}^t\log p(x_i|x_{i-1})\nonumber\\
    & = -\frac{1}{2\gamma_0}(x_1-\mu)^2-\frac{1}{2}\log \gamma_0-\frac{1}{2}\log 2\pi\nonumber\\
    & -\frac{1}{2(\gamma_0-\frac{\gamma_1^2}{\gamma_0})}\sum_{i=2}^t (x_i-\theta_R-\frac{\gamma_1}{\gamma_0}(x_{i-1}-\theta_R))^2-\frac{t-1}{2}\log (\gamma_0-\frac{\gamma_1^2}{\gamma_0})-\frac{t-1}{2}\log 2\pi.
\end{align}
Then the derivative of the log-likelihood with respect to $\theta_R$ is:
\begin{equation}
    \frac{1}{\gamma_0}(x_1-\theta_R)+\frac{1-\frac{\gamma_1}{\gamma_0}}{2(\gamma_0-\frac{\gamma_1^2}{\gamma_0})}\sum_{i=2}^t 2(x_i-\theta_R-\frac{\gamma_1}{\gamma_0}(x_{i-1}-\theta_R))
\end{equation}
thus the maximum likelihood solution for the parameter $\theta_R$ to the dependent case of order $q=1$ is:
\begin{equation}
    \theta_\text{ML} =\frac{x_1+\frac{1-\frac{\gamma_1}{\gamma_0}}{1-(\frac{\gamma_1}{\gamma_0})^2}[(1-\frac{\gamma_1}{\gamma_0})\sum_{i=2}^{t-1}x_i+x_t-\frac{\gamma_1}{\gamma_0} x_1]}{1+\frac{(1-\frac{\gamma_1}{\gamma_0})^2(t-1)}{1-(\frac{\gamma_1}{\gamma_0})^2}}.
\end{equation}
Let $p(\theta_R) = \mathcal{N}(\mu_0,\sigma_0^2)$ being a conjugate prior then for the variance of the mean holds:
\begin{align}    \lim_{t\rightarrow\infty}\sigma_t^2 & = \lim_{t\rightarrow\infty}\Big(\frac{1}{\gamma_0}+\frac{(t-1)(1-\frac{\gamma_1}{\gamma_0})^2}{\gamma_0(1-(\frac{\gamma_1}{\gamma_0})^2)}+\frac{1}{\sigma_0^2}\Big)^{-1}=0
\end{align}
and for the mean of the mean:
\begin{align}\label{eq: step2 prop2}
    \mu_t & = \frac{b_t+\frac{\mu_0}{\sigma_0^2}}{a_t+\frac{1}{\sigma_0^2}}\nonumber\\
    & = \frac{\frac{x_1}{\gamma_0}+\frac{(1-\frac{\gamma_1}{\gamma_0})^2\sum_{i=2}^{t-1}x_i+(1-\frac{\gamma_1}{\gamma_0})(x_t-\frac{\gamma_1}{\gamma_0} x_1)}{\gamma_0(1-(\frac{\gamma_1}{\gamma_0})^2)}+\frac{\mu_0}{\sigma_0^2}}{\frac{1}{\gamma_0}+\frac{(t-1)(1-\frac{\gamma_1}{\gamma_0})^2}{\gamma_0(1-(\frac{\gamma_1}{\gamma_0})^2)}+\frac{1}{\sigma_0^2}}\nonumber\\
    & = \frac{x_1(1-(\frac{\gamma_1}{\gamma_0})^2)\sigma_0^2+\sigma_0^2\Big[(1-\frac{\gamma_1}{\gamma_0})^2\sum_{i=2}^{t-1}x_i+(1-\frac{\gamma_1}{\gamma_0})(x_t-\frac{\gamma_1}{\gamma_0} x_1)\Big]+\mu_0\gamma_0(1-(\frac{\gamma_1}{\gamma_0})^2)}{(1-(\frac{\gamma_1}{\gamma_0})^2)\sigma_0^2+(t-1)(1-\frac{\gamma_1}{\gamma_0})^2\sigma_0^2+\gamma_0(1-(\frac{\gamma_1}{\gamma_0})^2)}\nonumber\\
    & = \frac{\sigma_0^2\Big[1-(\frac{\gamma_1}{\gamma_0})^2+(t-1)(1-\frac{\gamma_1}{\gamma_0})^2\Big]\theta_\text{ML}+\mu_0\gamma_0(1-(\frac{\gamma_1}{\gamma_0})^2)}{(1-(\frac{\gamma_1}{\gamma_0})^2)\sigma_0^2+(t-1)(1-\frac{\gamma_1}{\gamma_0})^2\sigma_0^2+\gamma_0(1-(\frac{\gamma_1}{\gamma_0})^2)}\nonumber\\
    & = \frac{\frac{1-(\frac{\gamma_1}{\gamma_0})^2}{t-1}+(1-\frac{\gamma_1}{\gamma_0})^2}{\frac{1-(\frac{\gamma_1}{\gamma_0})^2}{t-1}+(1-\frac{\gamma_1}{\gamma_0})^2+\frac{\gamma_0(1-(\frac{\gamma_1}{\gamma_0})^2)}{\sigma^2_0(t-1)}}\theta_\text{ML}+\frac{\gamma_0(1-(\frac{\gamma_1}{\gamma_0})^2)}{(1-(\frac{\gamma_1}{\gamma_0})^2)\sigma_0^2+(t-1)(1-\frac{\gamma_1}{\gamma_0})^2\sigma_0^2+\gamma_0(1-(\frac{\gamma_1}{\gamma_0})^2)}\mu_0
\end{align}
hence,
\begin{equation}
    \lim_{t\rightarrow\infty} \mu_t = \theta_{ML}.
\end{equation}
Finally, from Eq. (\ref{eq: step2 prop2}) we observe that the posterior mean is a weighted average between the maximum likelihood solution $\theta_\text{ML}$ and the prior mean value $\mu_0$.
\newline

%==================================
\iffalse
\section{Score-Driven MBO(1)}\label{app: Score-Driven MBO(1)}
In algorithm \ref{alg: MBOC V1} we show a slightly different version of algorithm \ref{alg: MBOC new}. In particular, the inference of the parameter vector is accomplished at each time, but within the most likely regime,  contrary to the case in algorithm \ref{alg: MBOC new}, where the inference is performed at each time, but using the entire dataset after we de-mean the past data given the most likely regimes.

\begin{algorithm}[tb]
\caption{MBOC V2}
\label{alg: MBOC V1}
\textbf{Input}: $\mu_0,\sigma_0^2,d,\vec\lambda_0, \rho_{1,d}, \sigma_i^2, p(r_0=0) = 1, \eta$\\
\textbf{Output}: $p(r_t|x_{1:t}), \hat{\mu}_t$
\begin{algorithmic}[1] %[1] enables line numbers
\For{$t = 1,...$} 
\State Observe $x_t\sim\mathcal{N}(\theta_{R},\sigma^2)$
\State Compute $\hat{\mu}_t$
\State Find $\text{argmax}_{i\in\{0,1,...,t\}}p(r_t=i|x_{1:t})$
\If{$t>1$ \textbf{\&} $i>\eta$}
\State Infer $\vec\lambda_t$ with GAS using $x_t^{(i)}-\frac{\sum_t x_t^{(i)}}{i} = \{x_{t-i+1}-\frac{\sum_t x_t^{(i)}}{i},\cdots,x_t-\frac{\sum_t x_t^{(i)}}{i}\}$
\State Filter $\rho_{t,d}$
\EndIf
\State Compute $p(r_t|x_{1:t})$\Comment{The correlation of the previous step $\rho_{t-1,d}$ is used here}
\State Update $\mu_t,\sigma_t^2$
\EndFor
\end{algorithmic}
\end{algorithm}
%===================================
\fi

\section{Sensitivity of the methods to the investigated period for the GNP growth rate}\label{app:emp}

\begin{figure}[t]
\centering
\includegraphics[width=1\columnwidth]{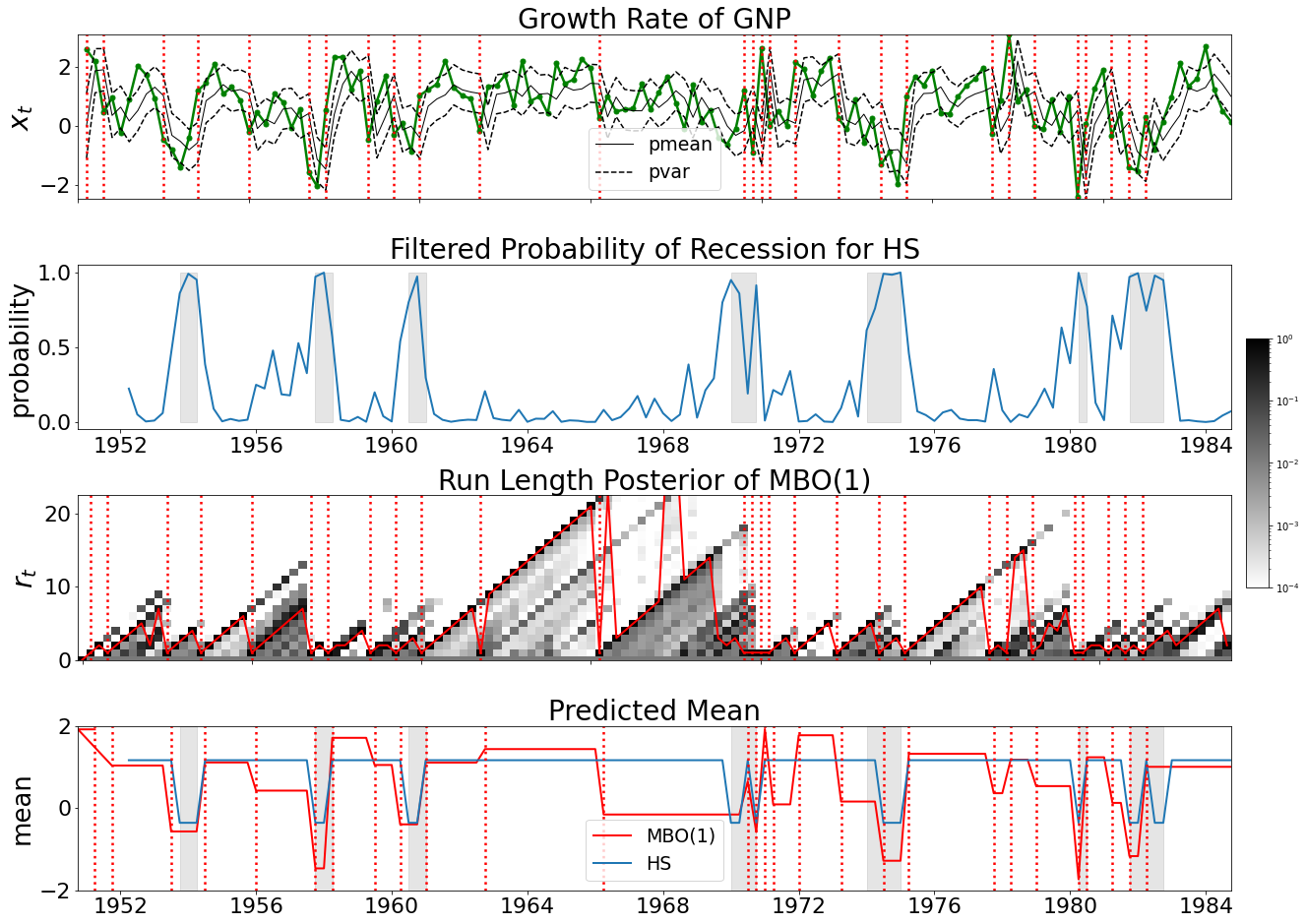}
\caption{\textbf{First:} The growth rate of U.S. GNP (green line) during 1951-1984. The red dotted lines represent the CPs found by MBO(1).
\textbf{Second:} The filtered probability of recession for HS model. The gray areas represent the periods of recession. The tick labels on the x-axis represent the first quarter of the indicated year. \textbf{Third:} The Run Length Posterior of MBO(1). The darkest
the color, the highest the probability of the run length. The red solid line represents the most likely path i.e. value of $r_t$ with the largest run length
posterior $p(r_t|x_{1:t})$ for each $t$. \textbf{Fourth:} The posterior mean of MBO(1) versus the predicted mean of HS model across the regimes identified by each method.}
\label{fig: GNP_MBO1_HS}
\end{figure}

\begin{figure}[t]
\centering
\includegraphics[width=1\columnwidth]{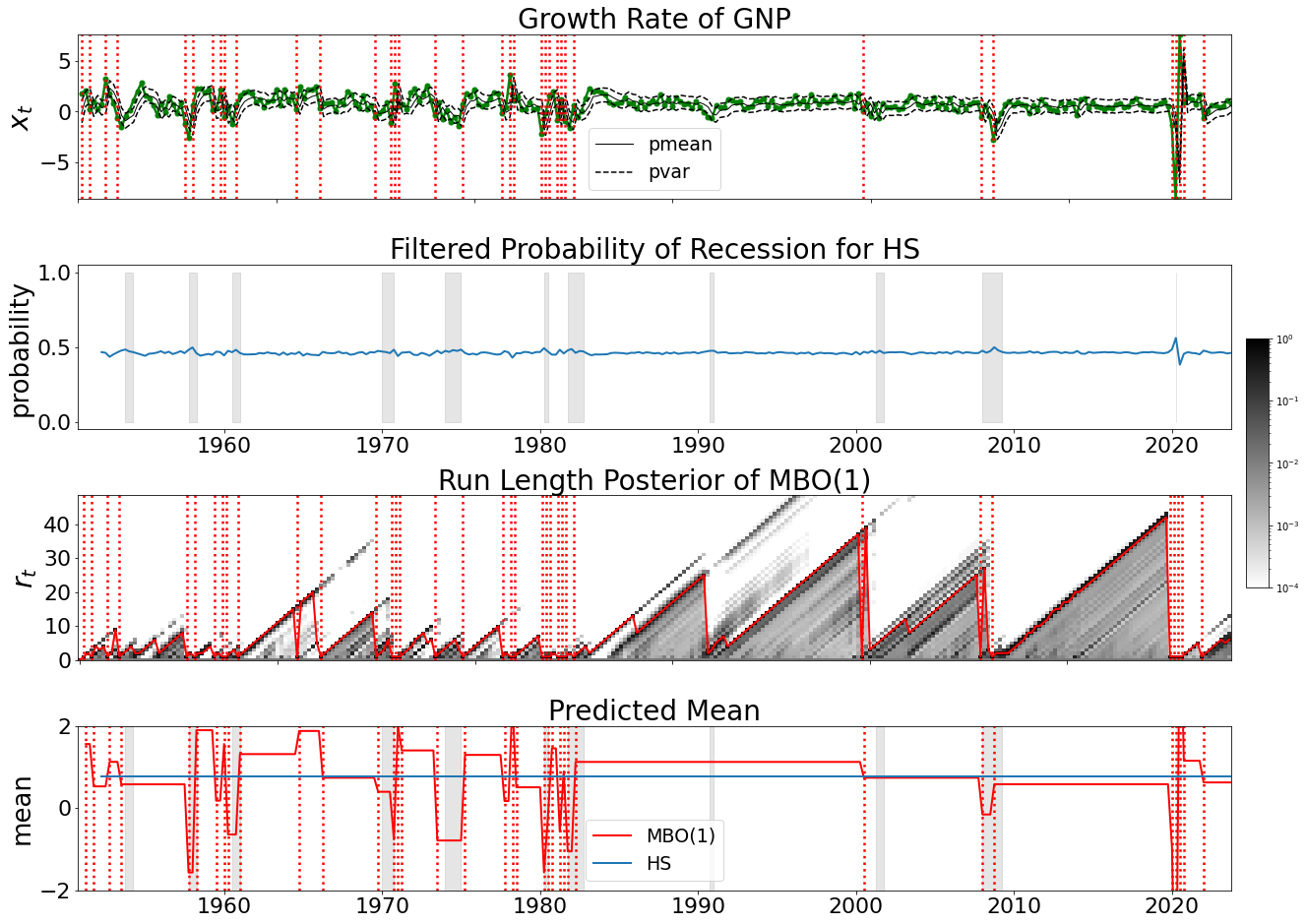}
\caption{\textbf{First:} The growth rate of U.S. GNP (green line) during $1951-2023$. The red dotted lines represent the CPs found by MBO(1). \textbf{Second:} The filtered probability of recession for HS model. The gray areas represent the periods of recession. The tick labels on the x-axis represent the first quarter of the indicated year. \textbf{Third:} The Run Length Posterior of MBO(1). The darkest the color, the highest the probability of the run length. The red solid line represents the most likely path i.e. value of $r_t$ with the largest run length posterior $p(r_t|x_{1:t})$ for each $t$. \textbf{Fourth:} The posterior mean of MBO(1) versus the predicted mean of HS model across the regimes identified by each method.}
\label{fig: GNP_MBO1_HS_V2}
\end{figure}

In this appendix, we compare the sensitivity of the HS model and of the MBO(1) to the investigated period. Differently from the latter, the HS model is estimated offline, so  regimes detected in the same period will not be the same if the overall estimation period is different. 

First of all, we investigate the same period (1951-1984) as in the original paper of \cite{Hamilton1989}. The top panel of Figure \ref{fig: GNP_MBO1_HS} shows the growth rate, the second the filtered probability of recession for the HS model, the third one the run length posterior of the MBO(1), and the fourth one the predicted mean for each model. We observe an overall agreement between the two methods, however since the HS assumes only two regimes, every recession is classified with the same mean even though this might not be true in reality. Differently from HS, the MBO(1) model that does not prespecify the number of regimes, is able to find different mean during the recessions but also during the periods of growth. Notice also that the period investigated in \cite{Hamilton1989} does not contain extreme events. 

On the contrary, when we consider the period
1951-2023 (see Figure \ref{fig: GNP_MBO1_HS_V2}) the economic turmoil around the Covid pandemic stands out as a very large event. Applying the HS model to this time span, we observe that the it performs quite badly, both in terms of the identified regimes and in terms of the different mean in the two regimes. MBO(1) model is instead much more robust and provides online estimation,  which are not dependent from extreme events in the data.

\end{document}